\documentclass[10pt,twocolumn,letterpaper]{article}

\usepackage{caption}
\usepackage{cvpr}
\usepackage{times}
\usepackage{epsfig}
\usepackage{graphicx}
\usepackage{amsmath}
\usepackage{amssymb}
\usepackage{multirow}
\usepackage{abstract}

\usepackage{booktabs}

\usepackage[dvipsnames]{xcolor}

\newcommand{\revisiondelete}[1]{}

\newcommand{\unit}[1]{\ensuremath{\,\mathrm{#1}}}

\usepackage[pagebackref=true,breaklinks=true,letterpaper=true,colorlinks,bookmarks=false]{hyperref}

\cvprfinalcopy 

\ifcvprfinal\fi
\begin{document}

\title{Fast and Flexible Indoor Scene Synthesis via\\Deep Convolutional Generative Models}

\author{Daniel Ritchie \thanks{Equal contribution}\\
Brown University
\and
Kai Wang \footnotemark[1]\\
Brown University
\and
Yu-an Lin\\
Brown University
}


\twocolumn[{
\renewcommand\twocolumn[1][]{#1}
\maketitle
\begin{center}
\renewcommand{\arraystretch}{0.1}
\begin{tabular}{c@{\hskip 1cm}c}
    \emph{Bedrooms} & \emph{Living Rooms}\\
    \setlength{\tabcolsep}{-3pt}
     \begin{tabular}{ccc}
        \includegraphics[width=0.17\linewidth]{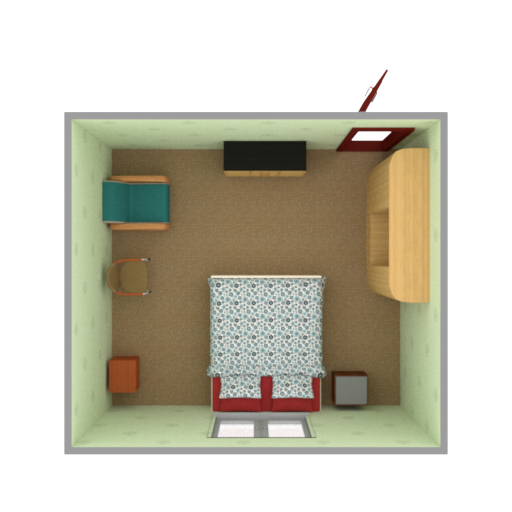} &
        \includegraphics[width=0.17\linewidth]{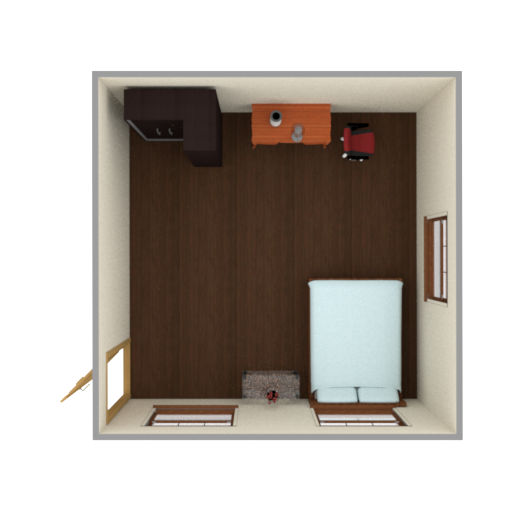} &
        \includegraphics[width=0.17\linewidth]{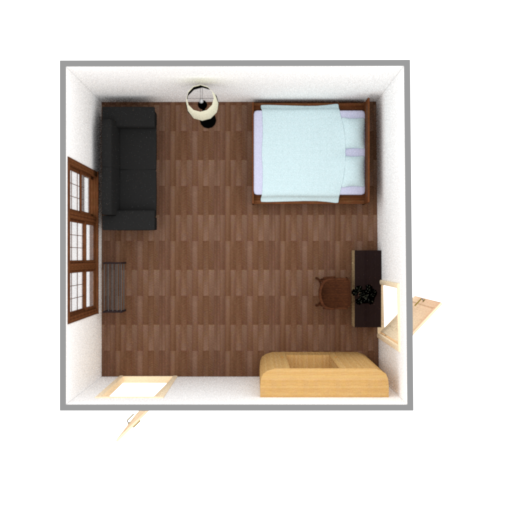}
     \end{tabular}
     &
     \setlength{\tabcolsep}{-3pt}
     \begin{tabular}{ccc}
        \includegraphics[width=0.17\linewidth]{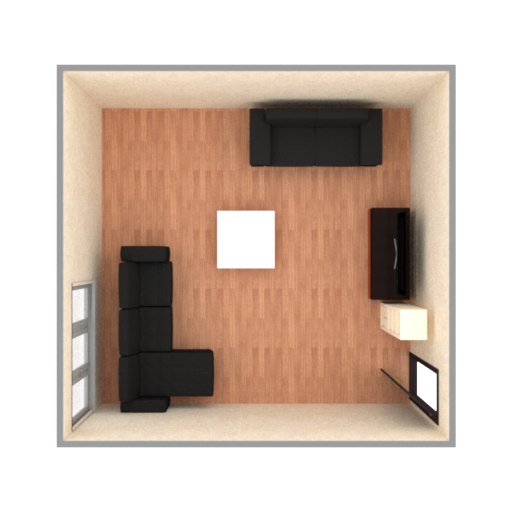} &
        \includegraphics[width=0.17\linewidth]{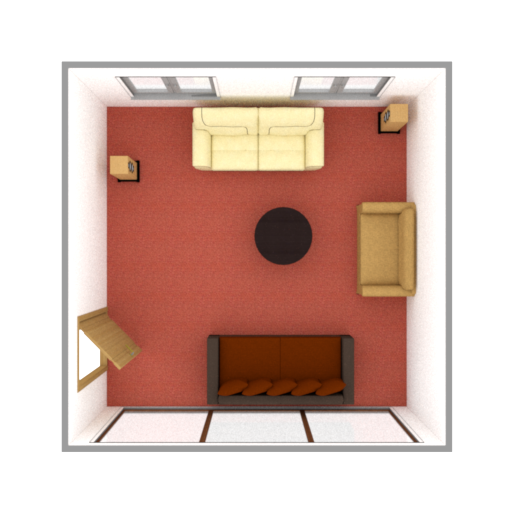} &
        \includegraphics[width=0.17\linewidth]{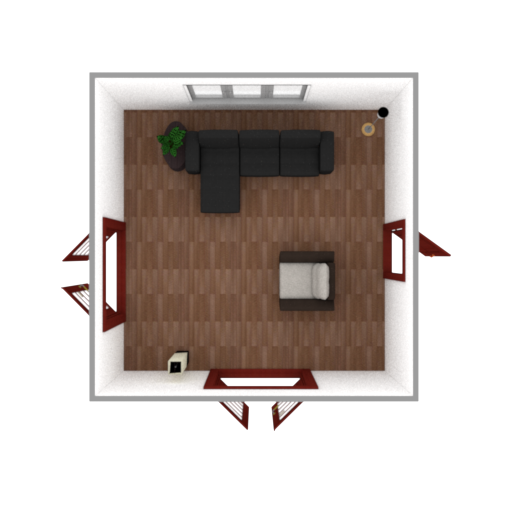}
     \end{tabular}
     \\
     \emph{Offices} & \emph{Bathrooms}\\
     \setlength{\tabcolsep}{-3pt}
     \begin{tabular}{ccc}
        \includegraphics[width=0.17\linewidth]{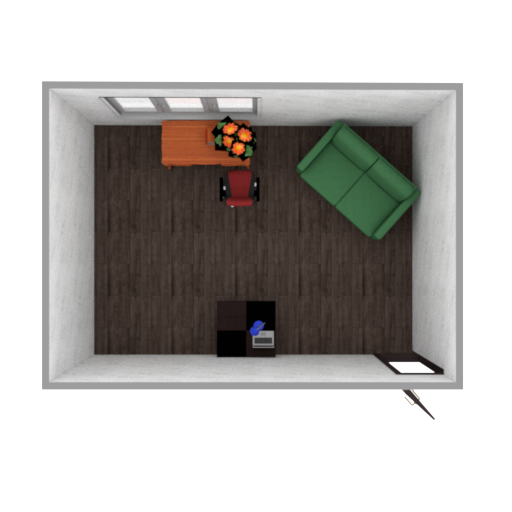} &
        \includegraphics[width=0.17\linewidth]{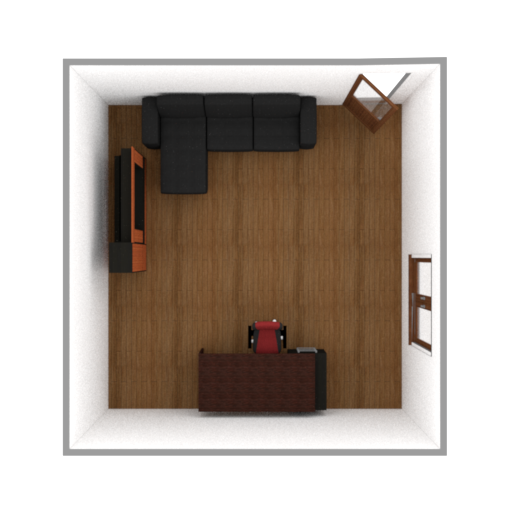} &
        \includegraphics[width=0.17\linewidth]{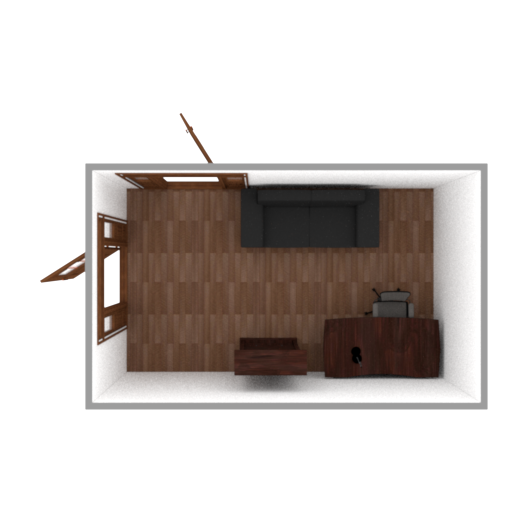}
     \end{tabular}
     &
     \setlength{\tabcolsep}{-3pt}
     \begin{tabular}{ccc}
        \includegraphics[width=0.17\linewidth]{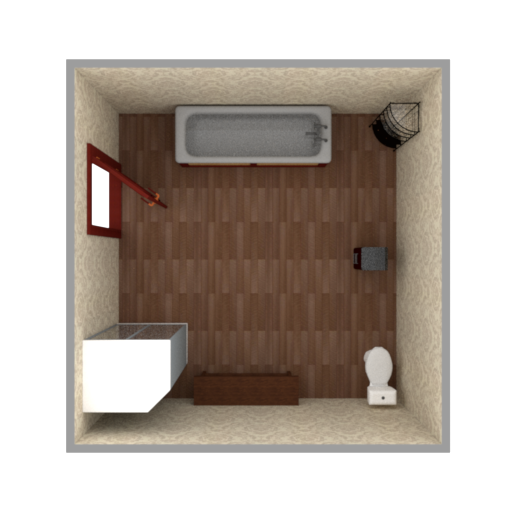} &
        \includegraphics[width=0.17\linewidth]{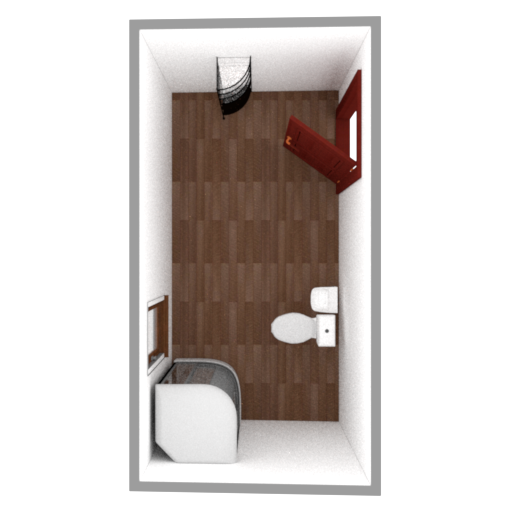} &
        \includegraphics[width=0.17\linewidth]{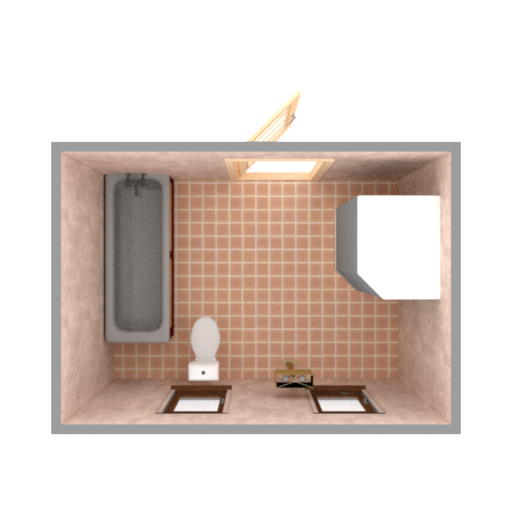}
     \end{tabular}
\end{tabular}
\end{center}
\captionof{figure}{Synthetic virtual scenes generated by our method. Our model can generate a large variety of such scenes, as well as complete partial scenes, in under two seconds per scene. This performance is enabled by a pipeline of multiple deep convolutional generative models which analyze a top-down representation of the scene.}
\label{fig:completesynth}
\vspace{0.5cm}
}]

\saythanks

\begin{abstract}
We present a new, fast and flexible pipeline for indoor scene synthesis that is based on deep convolutional generative models.
Our method operates on a top-down image-based representation, and inserts objects iteratively into the scene by predicting their category, location, orientation and size with separate neural network modules.
Our pipeline naturally supports automatic completion of partial scenes, as well as synthesis of complete scenes.
Our method is significantly faster than the previous image-based method and generates result that outperforms state-of-the-art generative scene models in terms of faithfulness to training data and perceived visual quality.

\end{abstract}

\section{Introduction}
\label{sec:intro}


People spend a large percentage of their lives indoors: in bedrooms, living rooms, offices, kitchens, and other such spaces.
The demand for virtual versions of these real-world spaces has never been higher.
Games, virtual reality, and augmented reality experience often take place in such environments.
Architects often create virtual instantiations of proposed buildings, which they visualize for customers using computer-generated renderings and walkthrough animations.
People who wish to redesign their living spaces can benefit from a growing array of available online virtual interior design tools~\cite{RoomSketcher,Planner5d}.
Furniture design companies, such as IKEA and Wayfair, increasingly produce marketing imagery by rendering virtual scenes, as it is faster, cheaper, and more flexible to do so than to stage real-world scenes~\cite{IkeaRendering}.
Finally, and perhaps most significantly, computer vision and robotics researchers have begun turning to virtual environments to train data-hungry models for scene understanding and autonomous navigation~\cite{ScanComplete,EmbodiedQA,IQA}. 

Given the recent interest in virtual indoor environments, a \emph{generative model} of interior spaces would be valuable.
Such a model would provide learning agents a strong prior over the structure and composition of 3D scenes.
It could also be used to automatically synthesize large-scale virtual training corpora for various vision and robotics tasks.


We define such a \emph{scene synthesis} model as an algorithm which, given an empty interior space delimited by architectural geometry (floor, walls, and ceiling), decides which objects to place in that space and where to place them.
Any model which solves this problem must reason about the existence and spatial relationships between objects in order to make such decisions.
In computer vision, the most flexible, general-purpose mechanism available for such reasoning is convolution, especially as realized in the form of deep convolutional neural networks (CNNs) for image understanding.
Recent work has attempted to perform scene synthesis using deep CNNs to construct priors over possible object placements in scenes~\cite{DeepSynthSIGGRAPH2018}.
While promising, this first attempt suffers from many limitations.
It reasons locally about object placements and can struggle to globally coordinate an entire scene (e.g. failing to put a sofa into a living room scene).
It does not model the size of objects, leading to problems with inappropriate object selection (e.g. an implausibly-long wardrobe which blocks a doorway).
Finally, and most critically, it is extremely slow, requiring minutes to synthesize a scene due to its use of hundreds of deep CNN evaluations per scene.

We believe that image-based synthesis of scenes is promising because of the ability to perform precise, pixel-level spatial reasoning, as well as the potential to leverage existing sophisticated machinery developed for image understanding with deep CNNs.
In this paper, we present a new image-based scene synthesis pipeline, based on deep convolutional generative models, that overcomes the issues of prior image-based synthesis work.
Like the previous method mentioned above, it generates scenes by iteratively adding objects.
However, it factorizes the step of adding each object into a different sequence of decisions which allow it (a) to reason globally about which objects to add, and (b) to model the spatial extent of objects to be added, in addition to their location and orientation.
Most importantly, it is fast: two orders of magnitude faster than prior work, requiring on average under 2 seconds to synthesize a scene.

We evaluate our method by using it to generate synthetic bedrooms, living rooms, offices, and bathrooms (Figure~\ref{fig:completesynth}).
We also show how, with almost no modification to the pipeline, our method can synthesize multiple automatic completions of partial scenes using the same fast generative procedure.
We compare our method to the prior image-based method, another state-of-the art deep generative model based on scene hierarchies, and scenes created by humans, in several quantitative experiments and a perceptual study.
Our method performs as well or better than these prior techniques.
\section{Related Work}
\label{sec:relatedwork}

\paragraph{Indoor Scene Synthesis}
A considerable amount of effort has been devoted to studying indoor scene synthesis. Some of the earliest work in this area utilizes interior design principles~\cite{InteractiveFurnitureLayout} and simple statistical relationships~\cite{MakeItHome} to arrange pre-specified sets of objects. Other early work attempts fully data-driven scene synthesis~\cite{SceneSynth} but is limited to small scale scenes due to the limited availability of training data and the learning methods available at the time.

With the availability of large scene datasets such as SUNCG~\cite{SUNCG}, new data-driven methods have been proposed.
\cite{EdinburghConstrainedSynth} uses a directed graphical model for object selection but relies on heuristics for object layout.
\cite{HumanCentricSUNCGSceneSynth} uses a probabilistic grammar to model scenes, but also requires data about human activity in scenes (not readily available in all datasets) as well as manual annotation of important object groups. In contrast, our model uses deep convolutional generative models to generate all important object attributes---category, location, orientation and size---fully automatically. 

Other recent methods have adapted deep neural networks for scene synthesis.
\cite{QixingSynth} uses a Generative Adversarial Network to generate scenes in an attribute-matrix form (i.e. one column per scene object).
More recently, GRAINS~\cite{GRAINS} uses recursive neural networks to encode and sample structured scene hierarchies. 
Most relevant to our work is~\cite{DeepSynthSIGGRAPH2018}, which also uses deep convolutional neural networks that operate on top-down image representations of scenes and synthesizes scenes by sequentially placing objects.
The main difference between our method and theirs is that (1) it sample each object attribute with a single inference step, while theirs perform hundreds of inferences, and (2) our method models the distribution over object size in addition to category, location, and orientation.
Our method also uses separate modules to predict category and location, thus avoiding some of the failure cases their method exhibits.

\begin{figure*}[ht!]
    \includegraphics[width=\textwidth]{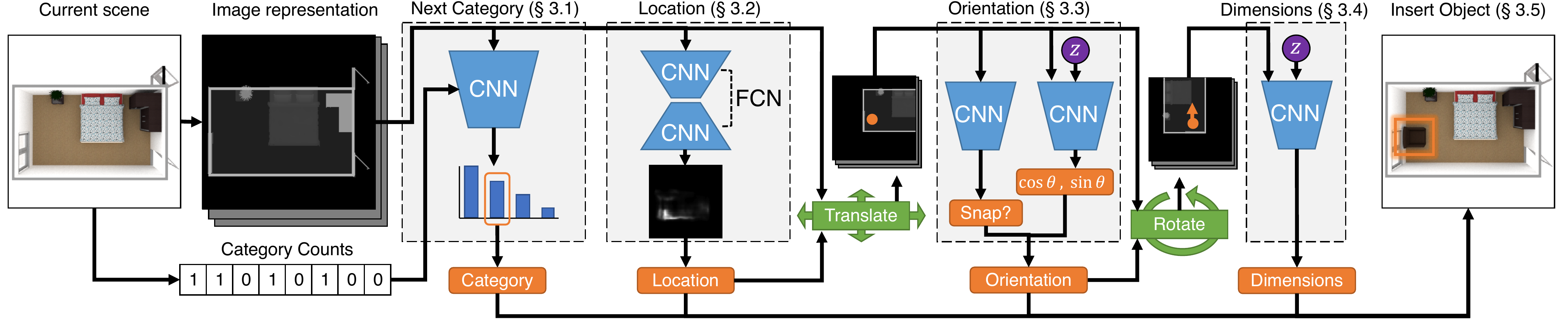}
\caption{
Overview of our automatic object-insertion pipeline.
We extract a top-down-image-based representation of the scene, which is fed to four decision modules:
which category of object to add (if any), the location, orientation, and dimensions of the object.
}
\label{fig:pipeline}
\end{figure*}

\paragraph{Deep Generative Models}
Deep neural networks are increasingly used to build powerful models which \emph{generate} data distributions, in addition to analyzing them, and our model leverages this capability.
Deep latent variable models, in particular variational autoencoders (VAEs)~\cite{VAE} and generative adversarial networks (GANs)~\cite{GAN}, are popular for their ability to pack seemingly arbitrary data distributions into well-behaved, lower-dimensional ``latent spaces.''
Our model uses conditional variants of these models---CVAEs~\cite{CVAE} and CGANs~\cite{CGAN}---to model the potentially multimodal distribution over object orientation and spatial extent.
Deep neural networks have also been effectively deployed for decomposing complex distributions into a sequence of simpler ones.
Such sequential or autoregressive generative models have been used for unsupervised parsing of objects in images~\cite{AIR}, generating natural images with sequential visual attention~\cite{DRAW}, parsing images of hand-drawn diagrams~\cite{TikzPaper}, generating 3D objects via sequential assemblies of primitives~\cite{3DPRNN}, and controlling the output of procedural graphics programs~\cite{NGPM}, among other applications.
We use an autoregressive model to generate indoor scenes, constructing them object by object, where each step is conditioned on the scene generated thus far.

\paragraph{Training Data from Virtual Indoor Scenes}
Virtual indoor scenes are rapidly becoming a crucial source of training data for computer vision and robotics systems.
Several recent works have shown that indoor scene understanding models can be improved by training on large amounts of synthetically-generated images from virtual indoor scenes~\cite{RenderingSUNCG}.
The same has been shown for indoor 3D reconstruction~\cite{ScanComplete}, as well as localization and mapping~\cite{InteriorNet}.
At the intersection of vision and robotics, researchers working on \emph{visual navigation} often rely on virtual indoor environments to train autonomous agents for tasks such as interactive/embodied question answering~\cite{EmbodiedQA,IQA}.
To support such tasks, a myriad of virtual indoor scene simulation platforms have emerged in recent years~\cite{MINOS,House3D,HoME,AI2Thor,CHALET,VirtualHome}.
Our model can complement these simulators by automatically generating new environments in which to train such intelligent visual reasoning agents.
\section{Model}
\label{sec:model}


Our goal is to build a deep generative model of scenes that leverages precise image-based reasoning, is fast, and can flexibly generate a variety of plausible object arrangements.
To maximize flexibility, we use a sequential generative model which iteratively inserts one object at a time until completion.
In addition to generating complete scenes from an empty room, this paradigm naturally supports partial scene completion by simply initializing the process with a partially-populated scene.
Figure~\ref{fig:pipeline} shows an overview of our pipeline.
It first extracts a top-down, floor-plan image representation of the input scene, as done in prior work on image-based scene synthesis~\cite{DeepSynthSIGGRAPH2018}.
Then, it feeds this representation to a sequence of four decision modules to determine how to select and add objects into the scene.
These modules decide which category of object to add to the scene, if any (Section~\ref{sec:nextcat}), where that object should be located (Section~\ref{sec:loc}), what direction it should face (Section~\ref{sec:orient}), and its physical dimensions (Section~\ref{sec:dims}).
This is a different factorization than in prior work, which we will show leads to both faster synthesis and higher-quality results.
The rest of this section describes the pipeline at a high level; precise architectural details can be found in Appendix~\ref{appendix:architecture}. 

\subsection{Next Object Category}
\label{sec:nextcat}

The goal of our pipeline's first module is, given a top down scene image representation, to predict the category of an object to add to the scene.
The module needs to reason about what objects are already present, how many, and the available space in the room.
To allow the model to also decide when to stop, we augment the category set with an extra ``$<$STOP$>$'' category.
The module uses a Resnet18~\cite{ResNet} to encode the scene image.
It also extract the counts of all categories of objects in the scene (i.e. a ``bag of categories'' representation), as in prior work~\cite{DeepSynthSIGGRAPH2018}, and encodes this with a fully-connected network.
Finally, the model concatenates these two encodings and feeds them through another fully-connected network to output a probability distribution over categories.
At test time, the module samples from the predicted distribution to select the next category.

\begin{figure}[t!]
\centering
\setlength{\tabcolsep}{2pt}
\begin{tabular}{cc}
    \includegraphics[width=0.28\linewidth]{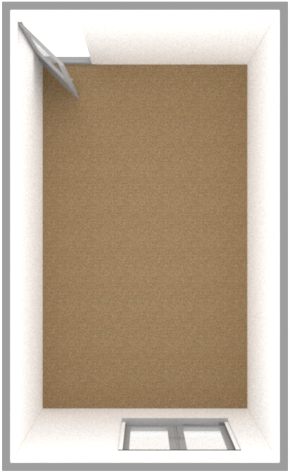} &
    \includegraphics[width=0.66\linewidth]{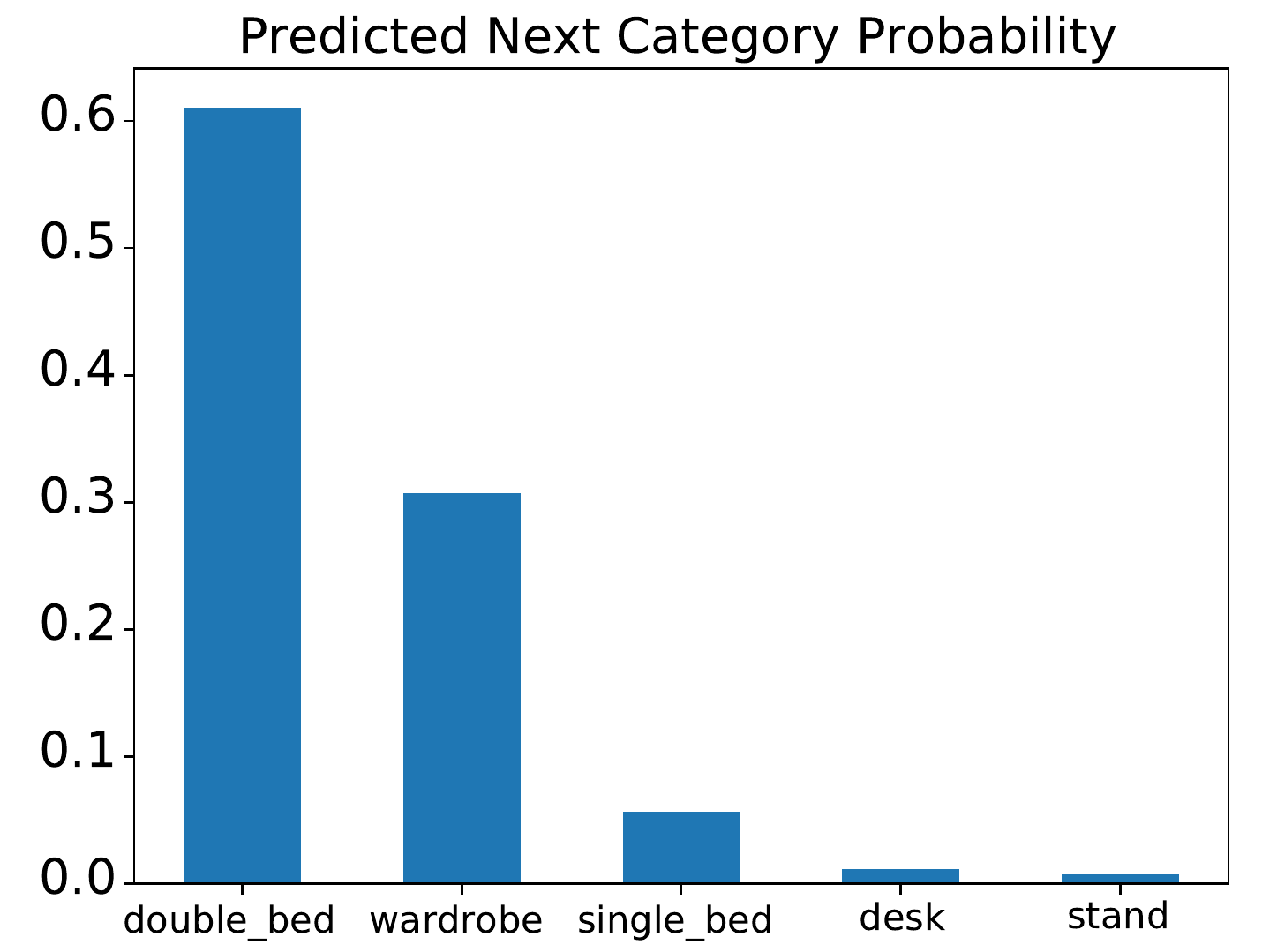}
    \\
    \includegraphics[width=0.28\linewidth]{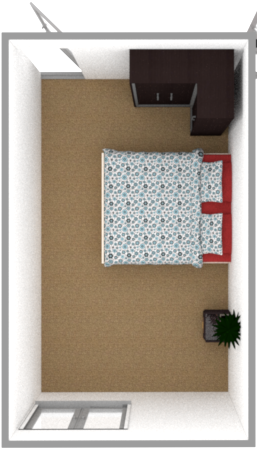} &
    \includegraphics[width=0.66\linewidth]{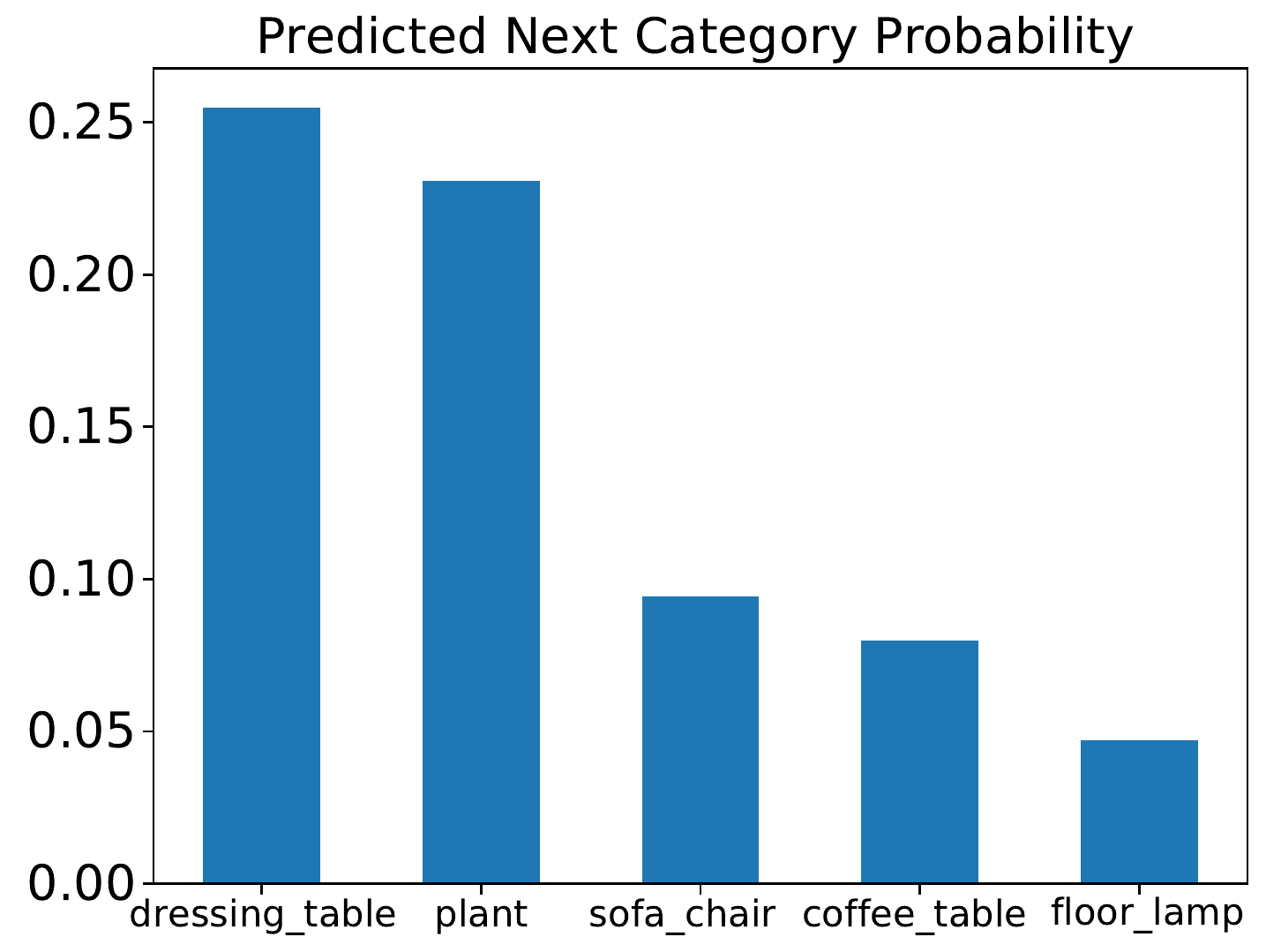}
    \\
    \raisebox{2.5em}{\includegraphics[width=0.33\linewidth]{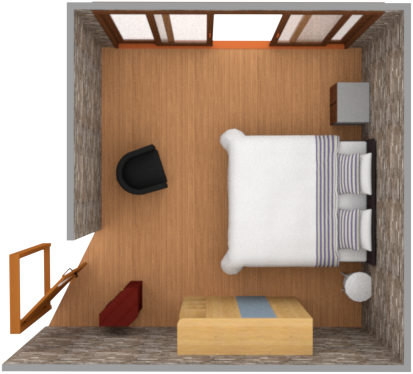}} &
    \includegraphics[width=0.66\linewidth]{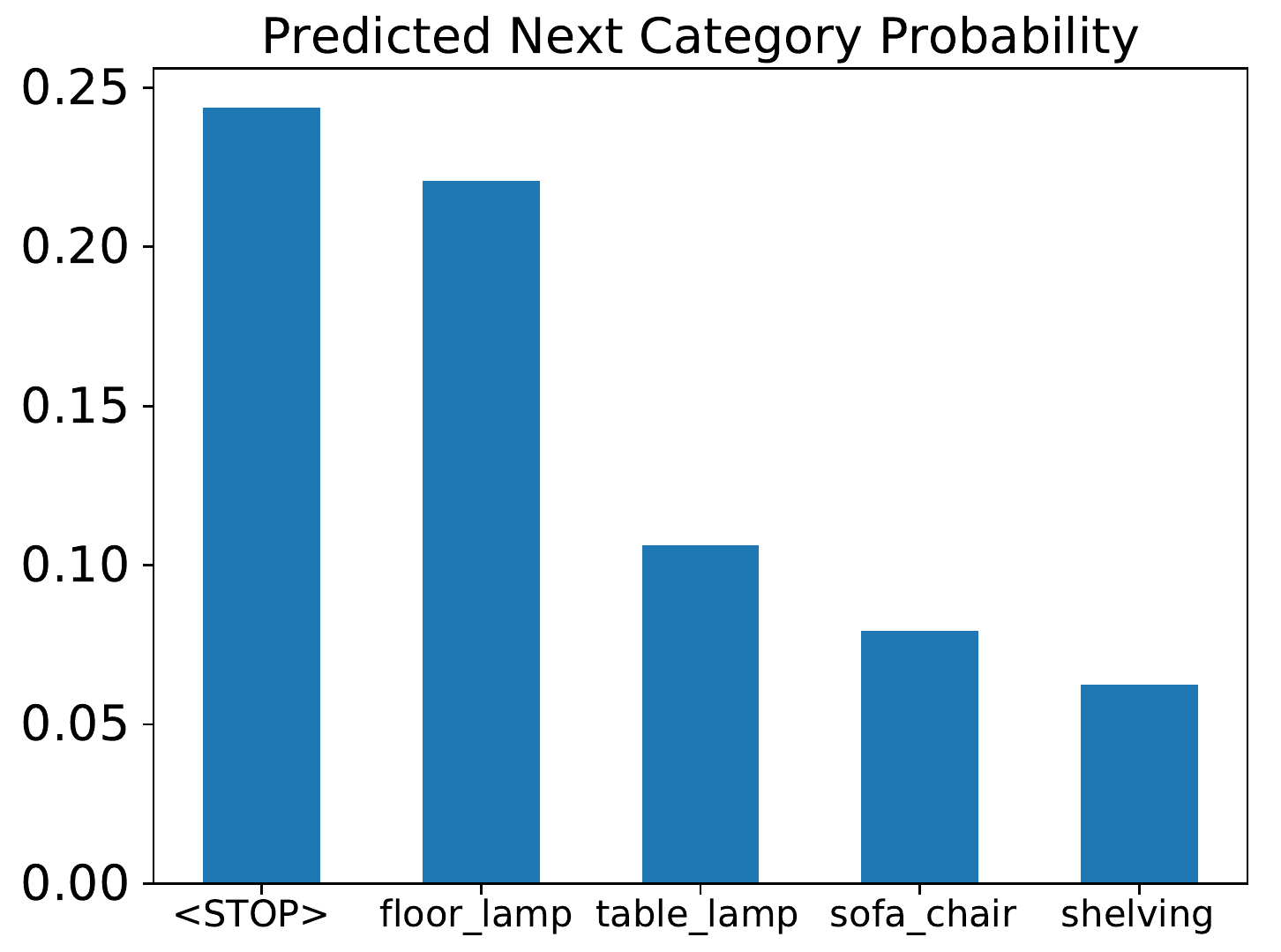}
\end{tabular}
\caption{Distributions over the next category of object to add to the scene, as predicted by our model. Empty scenes are dominated by one or two large, frequent object types (\emph{top}), partially populated scenes have a range of possibilities (\emph{middle}), and very full scenes are likely to stop adding objects (\emph{bottom}).}
\label{fig:nextcat}
\end{figure}

Figure~\ref{fig:nextcat} shows some example partial scenes and the most probable next categories that our model predicts for them.
Starting with an empty scene, the next-category distribution is dominated by one or two large, frequently-occurring objects (e.g. beds and wardrobes, for bedroom scenes).
The probability of other categories increases as the scene begins to fill, until the scene becomes sufficiently populated and the ``$<$STOP$>$'' category begins to dominate.

Prior work in image-based scene synthesis predicted category and location jointly~\cite{DeepSynthSIGGRAPH2018}.
This lead to the drawback, as the authors has noted, that objects which are very likely to occur in a location can be repeatedly (i.e. erroneously) sampled, e.g. placing multiple nightstands to the left of a bed.
In contrast, our category prediction module reasons about the scene globally and thus avoid this problem.

\subsection{Object Location}
\label{sec:loc}

\begin{figure}[t!]
\centering
\setlength{\tabcolsep}{2.5pt}
\begin{tabular}{ccc}
    \includegraphics[width=0.33\linewidth]{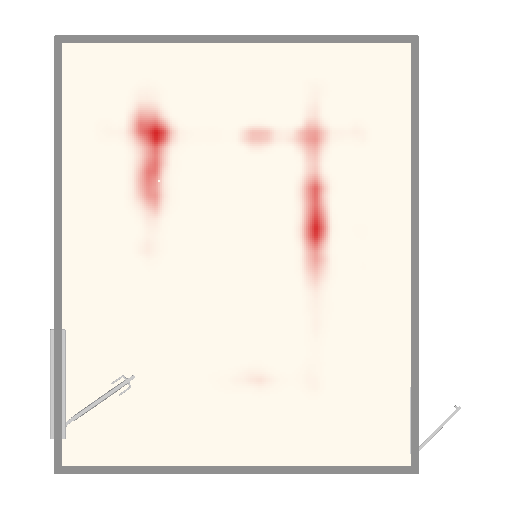} &
    \includegraphics[width=0.33\linewidth]{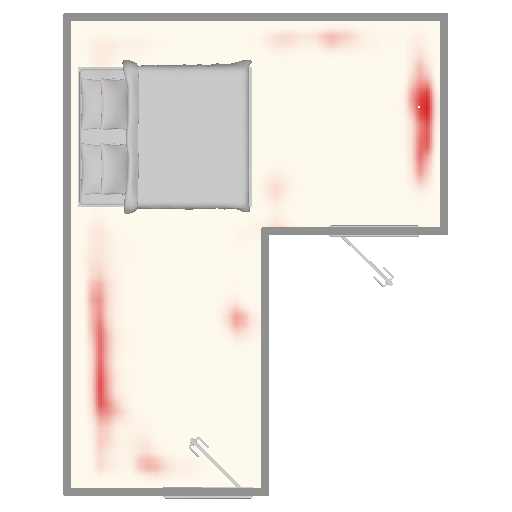} &
    \includegraphics[width=0.33\linewidth]{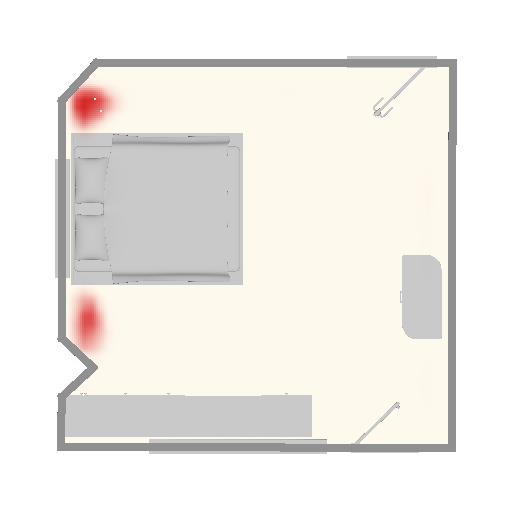}
    \\
    \emph{Double Bed} & \emph{Wardrobe} & \emph{Nightstand}
\end{tabular}
\caption{Probability densities for the locations of different object types predicted by our fully-convolutional network module.}
\label{fig:location}
\end{figure}

In the next module, our model takes the input scene and predicted category to determine where in the scene an instance of that category should be placed.
We treat this problem as an image-to-image translation problem: given the input top-down scene image, output a `heatmap' image containing the probability per pixel of an object occurring there.
This representation is advantageous because it can be treated as a (potentially highly multimodal) 2D discrete distribution, which we can sample to produce a new location.
This pixelwise discrete distribution is similar to that of prior work, except they assembled the distribution pixel-by-pixel, invoking a deep convolutional network once per pixel of the scene~\cite{DeepSynthSIGGRAPH2018}.
In contrast, our module uses a single forward pass through a fully-convolutional encoder-decoder network (FCN) to predict the entire distribution at once.

This module uses a Resnet34 encoder followed by an up-convolutional decoder. 
The decoder outputs a $64 \times 64 \times |C|$ image, where $|C|$ is the number of categories.
The module then slices out the channel corresponding to the category of interest and treats it as a 2D probability distribution by renormalizing it. 
We also experimented with using separate FCNs per category that predict a $64 \times 64 \times 1$ probability density image but found it not to work as well.
We suspect that training the same network to predict all categories provides the network with more context about different locations,  e.g. instead of just learning that it should not predict a wardrobe at a location, it can also learn that this is because a nightstand is more likely to appear there.
Before renormalization, the module zeros out any probability mass that falls outside the bounds of the room.
When predicting locations for second-tier categories (e.g. table lamps), it also zeros out probability mass that falls on top of an object that was not observed as a supporting surface for that category in the dataset.
At test time, we sample from a tempered version of this discrete distribution (we use temperature $\tau = 0.8$ for all experiments in this paper).

To train the network, we use pixel-wise cross entropy loss.
As in prior work, we augment the category set with a category for ``empty space,'' which allows the network to reason about where objects should \emph{not} be, in addition to where they should.
Empty-space pixels are weighted $10$ times less heavily than occupied pixels in the training loss computation.
A small amount of L2 regularization and dropout is used to prevent overfitting.

Figure~\ref{fig:location} shows examples of predicted location distributions for different scenes.
The predicted distributions for bed and wardrobe avoid placing probability mass in locations which would block the doors.
The distribution for nightstand is bimodal, with each mode tightly concentrated around the head of the bed.

Before moving on to the next module, our system translates the input scene image so that it is centered about the predicted location.
This makes the subsequent modules translation-invariant.

\subsection{Object Orientation}
\label{sec:orient}

\begin{figure}[t!]
\centering
\setlength{\tabcolsep}{2.5pt}
\begin{tabular}{ccc}
    \frame{\includegraphics[width=0.33\linewidth]{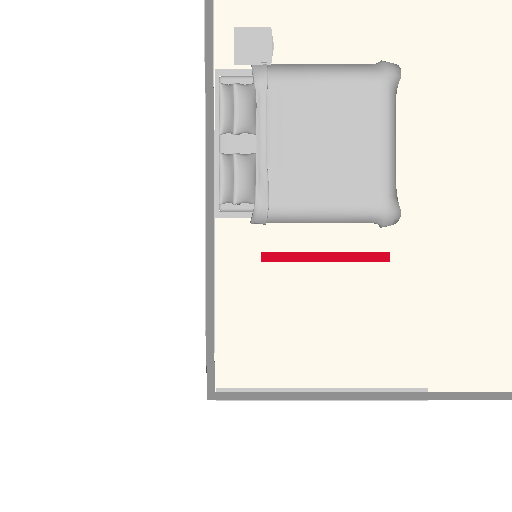}} &
    \frame{\includegraphics[width=0.33\linewidth]{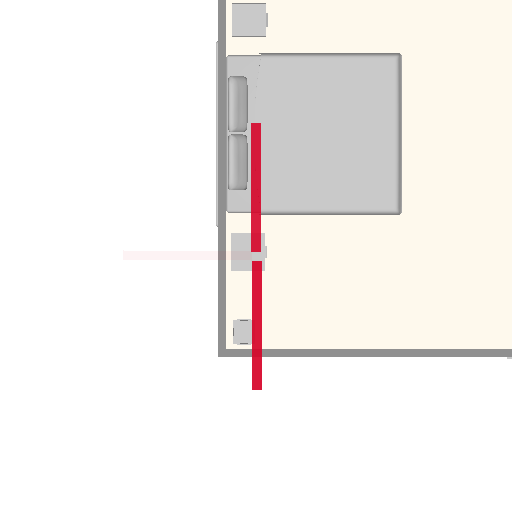}} &
    \frame{\includegraphics[width=0.33\linewidth]{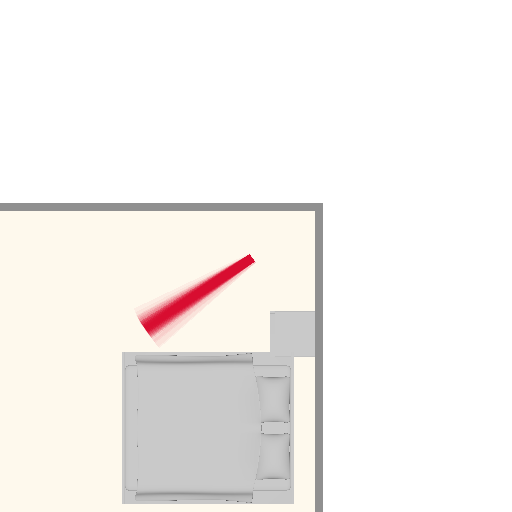}}
    \\
    \emph{Nightstand} & \emph{Table Lamp} & \emph{Armchair}
\end{tabular}
\caption{High-probability object orientations sampled by our CVAE orientation predictor (visualized as a density plot of front-facing vectors). Objects typically either snap to one orientation (\emph{left}) or multiple orientation modes (\emph{middle}), or have a range of values clustered around a single mode (\emph{right}).}
\label{fig:orient}
\end{figure}

Given a translated top-down scene image and object category, the orientation module predicts what direction an object of that category should face if placed at the center of the image.
We assume each category has a canonical front-facing direction.
Rather than predict the angle of rotation $\theta$, which is circular, we instead predict the front direction vector, i.e. $[\cos \theta, \sin \theta]$.
This must be a normalized vector, i.e. the magnitude of $\sin \theta$ must be $\sqrt{1 - \cos^2 \theta}$.
Thus, our module predicts $\cos \theta$ along with a Boolean value giving the sign of $\sin \theta$.
Here, we found using separate network weights per category to be most effective.

The set of possible orientations has the potential to be multimodal: for instance, a bed in the corner of a room may be backed up against either wall of the corner.
To allow our module to model this behavior, we implement it with a conditional variational autoencoder (CVAE)~\cite{CVAE}.
Specifically, we use a CNN to encode the input scene, which we then concatenate with a latent code $z$ sampled from a multivariate unit normal distribution, and then feed to a fully-connected decoder to produce $\cos \theta$ and the sign of $\sin \theta$.
At training time, we use the standard CVAE loss formulation (i.e. with an extra encoder network) to learn an approximate posterior distribution over latent codes).

Since interior scenes are frequently enclosed by rectilinear architecture, objects in them are often precisely aligned to cardinal directions.
A CVAE, however, being a probabilistic model, samples noisy directions.
To allow our module to produce precise alignments when appropriate, this module includes a second CNN which takes the input scene and predicts whether the object to be inserted should have its predicted orientation ``snapped'' to the nearest of the four cardinal directions.

Figure~\ref{fig:orient} shows examples of predicted orientation distributions for different input scenes.
The nightstand snaps to a single orientation, being highly constrained by its relations to the bed and wall.
Table lamps are often symmetric, which leads to a predicted orientation distribution with multiple modes.
An armchair to be placed in the corner of a room is most naturally oriented diagonally with respect to the corner, but some variability is possible.

Before moving on to the next module, our system rotates the input scene image by the predicted angle of rotation.
This transforms the image into the local coordinate frame of the object category to be inserted, making subsequent modules rotation-invariant (in addition to already being translation-invariant).

\subsection{Object Dimensions}
\label{sec:dims}

\begin{figure}[t!]
\centering
\setlength{\tabcolsep}{2.5pt}
\begin{tabular}{ccc}
    \frame{\includegraphics[width=0.33\linewidth]{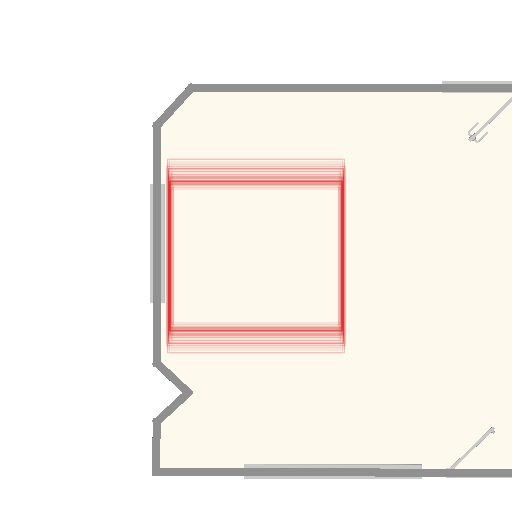}} &
    \frame{\includegraphics[width=0.33\linewidth]{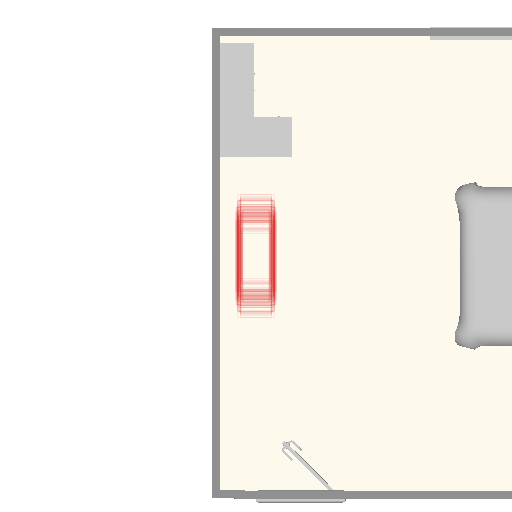}} &
    \frame{\includegraphics[width=0.33\linewidth]{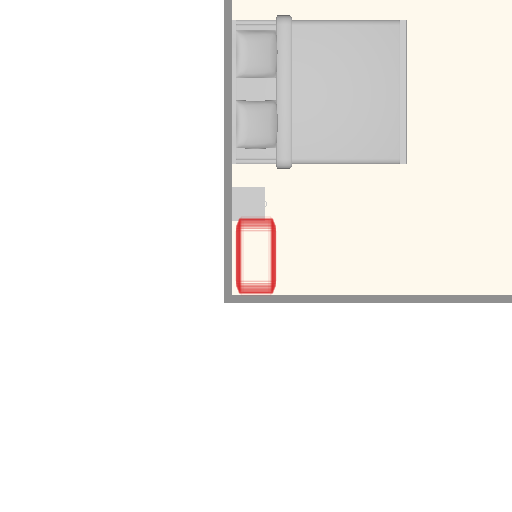}}
    \\
    \emph{Double Bed} & \emph{TV Stand} & \emph{Ottoman}
\end{tabular}
\caption{High-probability object dimensions sampled by our CVAE-GAN dimension predictor (visualized as a density plot of bounding boxes). Objects in more constrained locations have lower-variance size distributions (\emph{right}).}
\label{fig:dims}
\end{figure}

Given a scene image transformed into the local coordinate frame of a particular object category, the dimensions module predicts the spatial extent of the object.
That is, it predicts an object-space bounding box for the object to be inserted.
This is also a multimodal problem, even more so than orientation (e.g. many wardrobes of varying lengths can fit against the same wall).
Again, we use a CVAE for this: a CNN encodes the scene, concatenates it with $z$, and then uses a fully-connected decoder to produce the $[x,y]$ dimensions of the bounding box.

The human eye is very sensitive to errors in size, e.g. an object that is too large and thus penetrates the wall next to it.
To help fine-tune the prediction results, we also include an adversarial loss term in the CVAE training.
This loss uses a convolutional discriminator which takes the input scene concatenated channel-wise with the signed distance field (SDF) of the predicted bounding box.
As with the orientation module, this module also uses separate network weights per category.

Figure~\ref{fig:dims} visualizes predicted size distributions for different object placement scenarios.
The predicted distributions capture the range of possible sizes for different object categories, e.g. TV stands can have highly variable length.
However, in a situation such as Figure~\ref{fig:dims} Right, where an ottoman is to be placed between the nightstand and the wall, the predicted distribution is lower-variance due to this highly constrained location.

\subsection{Object Insertion}
\label{sec:insertion}

To choose a specific 3D model to insert given the predicted category, location, orientation, and size, we perform a nearest neighbor search through our dataset to find 3D models that closely fit the predicted object dimensions.
When multiple likely candidate models exist, we favor ones that have frequently co-occurred in the dataset with other objects already in the room, as this slightly improves the visual style of the generated rooms (though it is far from a general solution to the problem of style-aware scene synthesis).
Occasionally, the inserted object collides with existing objects in the room, or, for second-tier objects, overhangs too much over its supporting surface.
In such scenarios, we choose another object of the same category.
In very rare situations (less than $1\%$), no possible insertions exist.
If this occurs, we resample a different category from the predicted category distribution and try again.

\section{Data \& Training}
\label{sec:datatraining}

We train our model using the SUNCG dataset, a collection of over forty thousand scenes designed by users of an online interior design tool~\cite{SUNCG}.
In this paper, we focus our experiments on four common room types: bedrooms, living rooms, bathrooms, and offices.
We extract rooms of these types from SUNCG, performing pre-processing to filter out uncommon object types, mislabeled rooms, etc.
After pre-processing, we obtained 6300 bedrooms (with 40 object categories), 1400 living rooms (35 categories), 6800 bathrooms (22 categories), and 1200 offices (36 categories).
Further details about our dataset and pre-processing procedures can be found in Appendix~\ref{appendix:dataset}.

To generate training data for all of our modules, we follow the same general procedure: take a scene from our dataset, remove some subset of objects from it, and task the module with predicting the `next' object to be added (i.e. one of the removed objects).
This process requires an ordering of the objects in each scene.
We infer static support relationships between objects (e.g. lamp supported by table) using simple geometric heuristics, and we guarantee that all supported objects come after their supporting parents in this ordering.
We further guarantee that all such supported `second-tier' objects come after all `first-tier' objects (i.e. those supported by the floor).
For the category prediction module, we further order objects based on their \emph{importance}, which we define to be the average size of a category multiplied by its frequency of occurrence in the dataset.
Doing so imposes a stable, canonical ordering on the objects in the scene; without such an ordering, we find that there are too many valid possible categories at each step, and our model struggles to build coherent scenes across multiple object insertions.
For all other modules, we use a randomized ordering. 
Finally, for the location module, the FCN is tasked with predicting not the location of a single next object, but rather the locations of \emph{all} missing objects removed from the training scene whose supporting surface is present in the partial scene.

We train each module in our pipeline separately for different room categories.
Empirically, we find that the category module performs best after seeing $\sim300,000$ training examples, and the location module performs best after $\sim1,000,000$ examples. As the problems that the orientation and dimension models are solving is more local, their behavior is more stable across different epochs. In practice, with use orientation modules trained with $\sim2,000,000$ examples and dimension modules trained with $\sim1,000,000$ examples.
\section{Results \& Evaluation}
\label{sec:eval}


\paragraph{Complete scene synthesis}
Figure~\ref{fig:completesynth} shows examples of complete scenes synthesized by our model, given the initial room geometry.
Our model captures multiple possible object arrangement patterns for each room type: bedrooms with desks vs. those with extra seating, living rooms for conversation vs. watching television, etc.

\begin{figure*}[t!]
\setlength{\tabcolsep}{-1pt}
\begin{tabular}{c@{\hskip 1cm}ccc}
     Input Partial Scene & \multicolumn{3}{c}{Synthesized Completions}
     \\
     \includegraphics[width=0.24\linewidth]{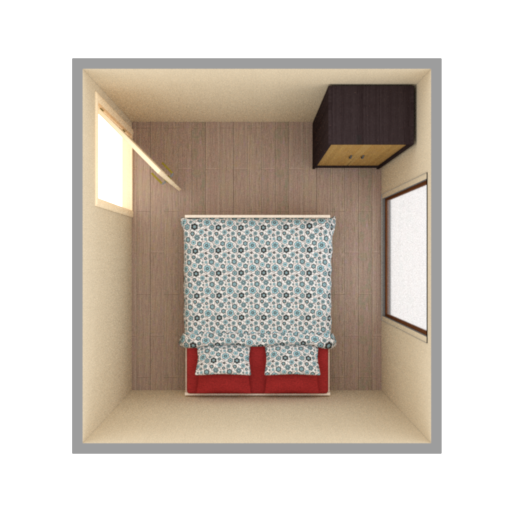} &
     \includegraphics[width=0.24\linewidth]{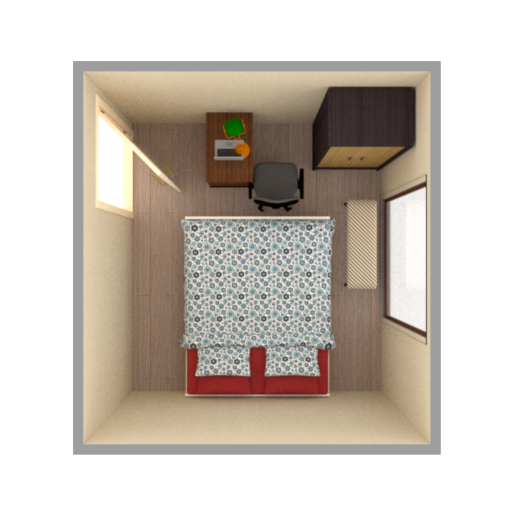}&
     \includegraphics[width=0.24\linewidth]{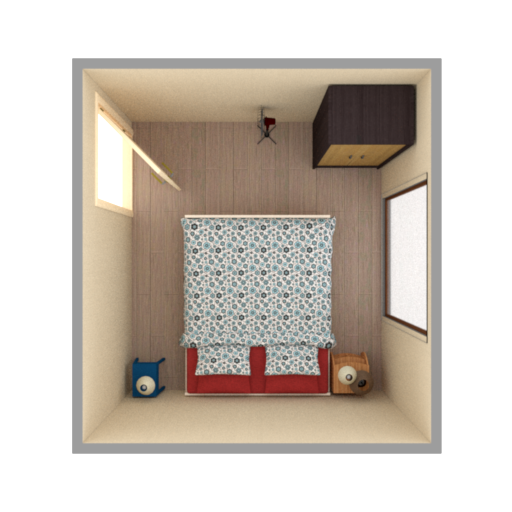}&
     \includegraphics[width=0.24\linewidth]{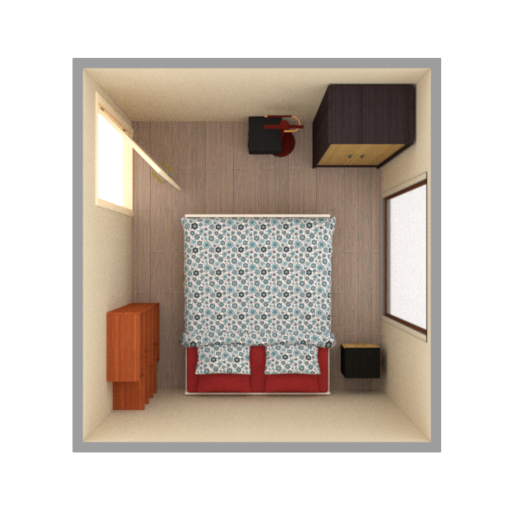}
     \\
     \addlinespace[-0.5cm]
     \includegraphics[width=0.24\linewidth]{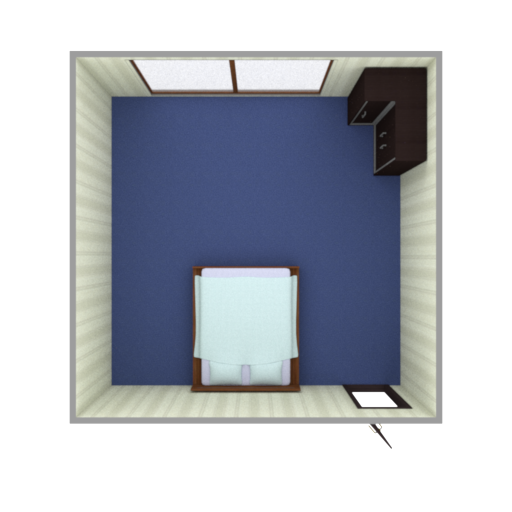} &
     \includegraphics[width=0.24\linewidth]{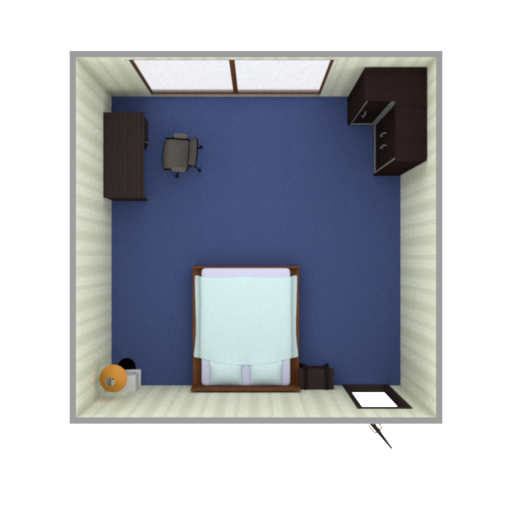}&
     \includegraphics[width=0.24\linewidth]{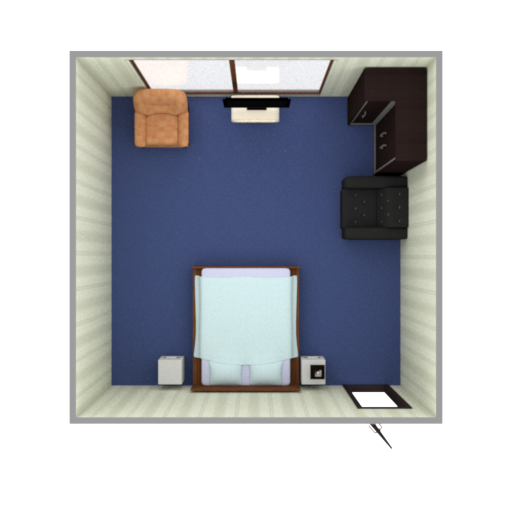}&
     \includegraphics[width=0.24\linewidth]{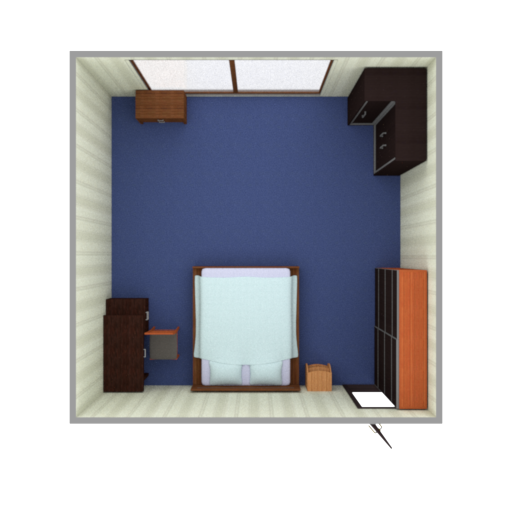}
     \\
     \addlinespace[-1.25cm]
     \includegraphics[width=0.24\linewidth]{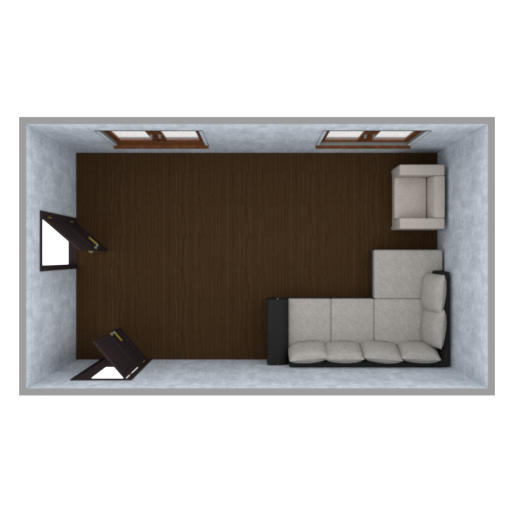} &
     \includegraphics[width=0.24\linewidth]{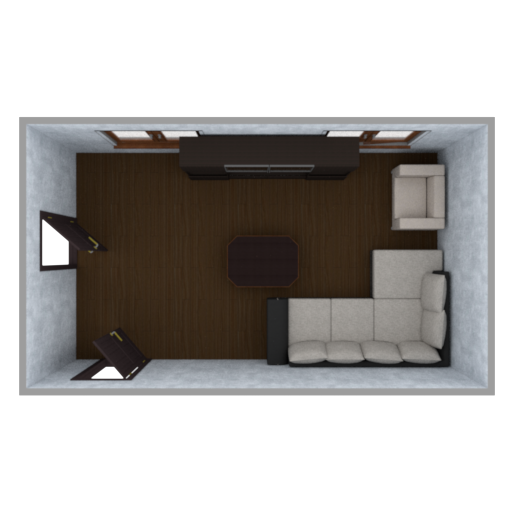} &
     \includegraphics[width=0.24\linewidth]{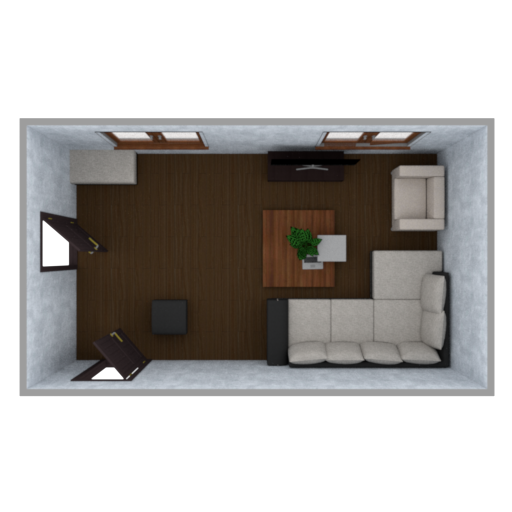} &
     \includegraphics[width=0.24\linewidth]{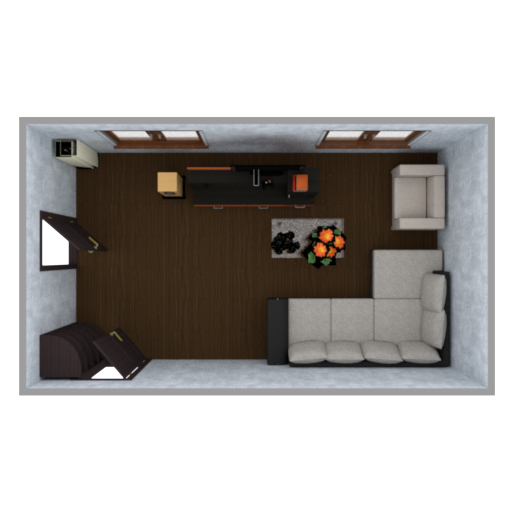}
     \\
     \addlinespace[-1.0cm]
     \includegraphics[width=0.24\linewidth]{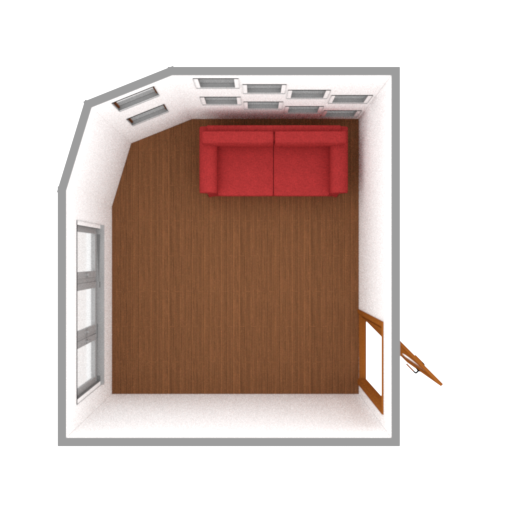} &
     \includegraphics[width=0.24\linewidth]{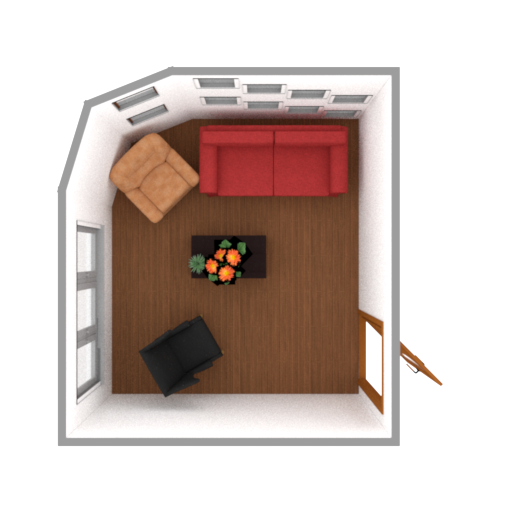} &
     \includegraphics[width=0.24\linewidth]{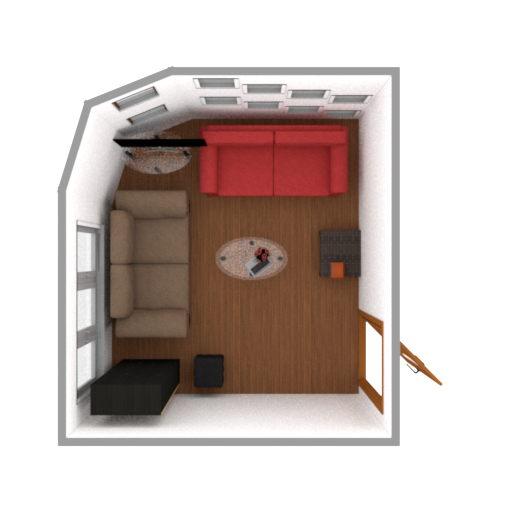} &
     \includegraphics[width=0.24\linewidth]{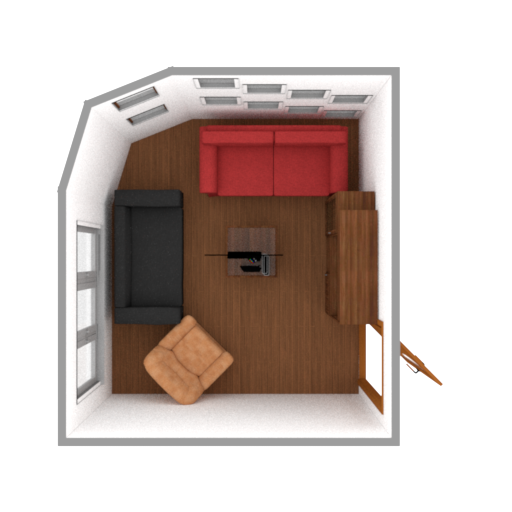}
\end{tabular}
    \caption{Given an input partial scene (\emph{left column}), our method can generate multiple automatic completions of the scene. This requires no modification to the method's sampling procedure, aside from seeding it with a partial scene instead of an empty one.}
    \label{fig:partialsynth}
\end{figure*}

\paragraph{Scene completion}
Figure~\ref{fig:partialsynth} shows examples of partial scene completion, where our model takes an incomplete scene as input and suggests multiple next objects to fill the scene.
Our model samples a variety of different completions for the same starting partial scene.
This example also highlights our model's ability to cope with non-rectangular rooms (bottom row), one of the distinct advantages of precise pixel-level reasoning with image-based models.

\paragraph{Object category distribution}
For a scene generative model to capture the training data well, a necessary condition is that the distribution of object categories which occurs in its synthesized results should closely resemble that of the training set.
To evaluate this, we compute the Kullback-Leibler divergence $D_{\text{KL}}(P_\text{synth}||P_\text{dataset})$ between the category distribution of synthesized scenes and that of the training set.
Note that we cannot compute a symmetrized Jensen–-Shannon divergence because some of the methods we compare against have zero probability for certain categories, making the divergence infinite.
Table~\ref{tab:catdists} shows the category distribution KL divergence of different methods.
Our method generates a category distribution that are more faithful to that of the training set than other approaches.

\begin{table}[t!]
    \centering
    \small
    \begin{tabular}{|c|c|c|c|c|}
        \hline
        \textbf{Method} & \emph{Bedroom} & \emph{Living} & \emph{Bathroom} & \emph{Office} \\
        \hline
        \emph{Uniform} & 0.6202 & 0.8858 & 1.3675 & 0.7219\\
        \hline
        \emph{Deep Priors~\cite{DeepSynthSIGGRAPH2018}} & 0.2017 & 0.4874 & 0.2479 & 0.2138 \\
        \hline
        \emph{Ours} & \textbf{0.0095} & \textbf{0.0179} & \textbf{0.0240} & \textbf{0.0436} \\
        \hline
    \end{tabular}
    \caption{KL divergence between the distribution of object categories in synthesized results vs. training set. Lower is better. \emph{Uniform} is the uniform distribution over object categories.}
    \label{tab:catdists}
\end{table}

\paragraph{Scene classification accuracy}
Looking beyond categories, to evaluate how well the distribution of our generated scenes match that of the training scenes, we train a classifier tasked to distinguish between ``real'' scenes (from the training set) and ``synthetic'' scenes (generated by our method).
The classifier is a Resnet34 that takes as input the same top-down multi-channel image representation that our model uses.
The classifier is trained with 1,600 scenes, half real and half synthetic.
We evaluate the classifier performance on 320 held out test scenes. 

Table~\ref{tab:classification_accuracy} shows the performance against different baselines. Compared to previous methods, our results are significantly harder for the classifier to distinguish.
In fact, it is marginally \emph{harder} to distinguish our scenes from real training scenes that it is to do so for scenes in which every object is perturbed by a small random amount (standard deviation of $10\%$ of the object's bounding box dimensions).

\begin{table}[t!]
    \centering
    \begin{tabular}{|c|c|}
        \hline
        \textbf{Method} & \textbf{Accuracy}\\
        \hline
        \emph{Deep Priors~\cite{DeepSynthSIGGRAPH2018}} & 84.69 \\
        \hline
        \emph{Human-Centric~\cite{HumanCentricSUNCGSceneSynth}} & 76.18 \\
        \hline
        \emph{Ours} & \textbf{58.75} \\
        \hline
        \hline
        \emph{Perturbed (1\%)} & 50.00 \\
        \hline
        \emph{Perturbed (5\%)} & 54.69 \\
        \hline
        \emph{Perturbed (10\%)} & 64.38 \\
        \hline
    \end{tabular}
    \caption{Real vs. synthetic classification accuracy for scenes generated by different methods. Lower (closer to 50\%) is better. Note that results of~\cite{HumanCentricSUNCGSceneSynth} are taken directly from their paper; their classification setup is largely similar but varies in details.}
    \label{tab:classification_accuracy}
\end{table}

\paragraph{Speed comparisons}
Table~\ref{tab:speed} shows the time taken for different methods to synthesize a complete scene. It takes on average less than $2$ seconds for our model to generate a complete scene on a NVIDIA Geforce GTX 1080Ti GPU, which is two orders of magnitudes faster than the previous image based method (Deep Priors). While slower than end-to-end methods such as \cite{GRAINS}, our model can also perform tasks such as scene completion and next object suggestion, both of which can be useful in real time applications.

\begin{table}[t!]
    \centering
    \begin{tabular}{|c|c|}
        \hline
        \textbf{Method} & \textbf{Avg. Time (\unit{s})}\\
        \hline
        \emph{Deep Priors~\cite{DeepSynthSIGGRAPH2018}} & $\sim 240$ \\
        \hline
        \emph{GRAINS~\cite{GRAINS}} & 0.1027 \\
        \hline
        \emph{Ours} & 1.858\\
        \hline
    \end{tabular}
    \caption{Average time in seconds to generate a single scene for different methods. Lower is better.}
    \label{tab:speed}
\end{table}



\paragraph{Perceptual study}
We also conducted a two-alternative forced choice (2AFC) perceptual study on Amazon Mechanical Turk to evaluate how plausible our generated scenes appear compared those generated by other methods.
Participants were shown two top-down rendered scene images side by side and asked to pick which one they found more plausible.
Images were rendered using solid colors for each object category, to factor out any effect of material or texture appearance.
For each comparison and each room type, we recruited 10 participants.
Each participant performed 55 comparisons; 5 of these were ``vigilance tests'' comparing against a randomly jumbled scene to check that participants were paying attention.
We filter out participants who did not pass all vigilance tests.

Table~\ref{tab:2afc} shows the results of this study.
Compared to the Deep Priors method, our scenes are preferred for bedrooms and bathrooms, and judged indistinguishable for living rooms.
Our generated office scenes are less preferred, however.
We hypothesize that this is because the office training data is highly multimodal, containing personal offices, group offices, conference rooms, etc.
It appears to us that the rooms generated by the Deep Priors method are mostly personal offices.
We also generate high quality personal offices consistently.
However, when the category module tries to sample other types of offices, this intent is not communicated well to other modules, resulting in unorganized results e.g. a small table with ten chairs.
Finally, compared to held-out human-created scenes from SUNCG, our results are indistinguishable for bedrooms and bathrooms, nearly indistinguishable for living rooms, and again less preferred for offices.

\begin{table}[t!]
    \small
    \centering
    \setlength{\tabcolsep}{2.5pt}
    \definecolor{gray}{gray}{0.5}
    \begin{tabular}{|c|c|c|c|c|}
        \hline
        \textbf{Ours vs.} & \emph{Bedroom} & \emph{Living} & \emph{Bathroom} & \emph{Office} \\
        \hline
        \emph{Deep Priors~\cite{DeepSynthSIGGRAPH2018}} & $\mathbf{56.1 \pm 4.1}$ & $52.7 \pm 4.5$ & $\mathbf{68.6 \pm 3.9}$ & $\textcolor{gray}{36.3 \pm 4.5}$ \\
        \hline
        \emph{SUNCG} & $48.0 \pm 4.7$ & $\textcolor{gray}{45.0 \pm 4.5}$ & $50.0 \pm 4.5$ & $\textcolor{gray}{34.8 \pm 5.1}$ \\
        \hline
    \end{tabular}
    \caption{Percentage ($\pm$ standard error) of forced-choice comparisons in which scenes generated by our method are judged as more plausible than scenes from another source. Higher is better. Bold indicate our scenes are preferred with $> 95\%$ confidence; gray indicates our scenes are dis-preferred with $> 95\%$ confidence; regular text indicates no preference.} 
    \label{tab:2afc}
\end{table}
\section{Conclusion}
\label{sec:conclusion}

In this paper, we presented a new pipeline for indoor scene synthesis using image-based deep convolutional generative models.
Our system analyzes top-down view representations of scenes to make decisions about which objects to add to a scene, where to add them, how they should be oriented, and how large they should be.
Combined, these decision modules allow for rapid (under 2 seconds) synthesis of a variety of plausible scenes, as well as automatic completion of existing partial scenes.
We evaluated our method via statistics of generated scenes, the ability of a classifier to detect synthetic scenes, and the preferences of people in a forced-choice perceptual study.
Our method outperforms prior techniques in all cases.

There are still many opportunities for future work in the area of automatic indoor scene synthesis.
We would like to address the limitations mentioned previously in our method's ability to generate room types with multiple strong modes of variation, e.g. single offices vs. conference offices.
One possible direction is to explore integrating our image-based models with models of higher-level scene structure, encoded as hierarchies a la GRAINS, or perhaps as graphs or programs.
Neither our method, nor any other prior work in automatic scene synthesis of which we are aware, addresses the problem of how to generate \emph{stylistically-consistent} indoor scenes, as would be required for interior design applications.
Finally, to make automatic scene synthesis maximally useful for training autonomous agents, generative models must be aware of the functionality of indoor spaces, and must synthesize environments that support carrying out activities of interest.


{\small
\bibliographystyle{ieee}
\bibliography{main}
}

%
%
%
%
%
%
%
%
%

%
%
\appendix

\section{Model Architecture Details}
\label{appendix:architecture}

\begin{figure*}[ht!]
    \includegraphics[width=\textwidth]{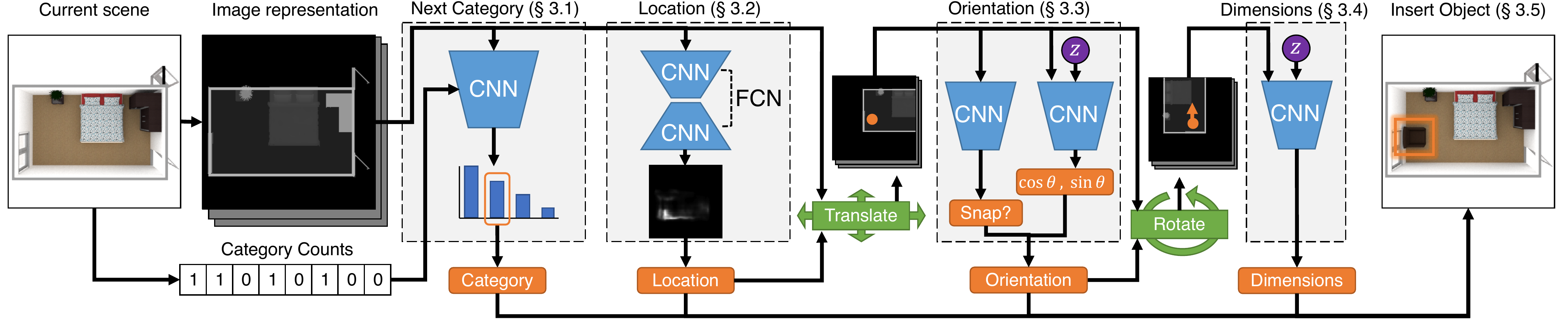}
\caption{
Overview of our automatic object-insertion pipeline.
We extract a top-down-image-based representation of the scene, which is fed to four decision modules:
which category of object to add (if any), the location, orientation, and dimensions of the object.
}
\label{fig:pipeline}
\end{figure*}

Here we give specific details about the neural network architectures used for each of our system's modules.
For reference, we also reproduce the pipeline overview figure from the main paper (Figure~\ref{fig:pipeline}).

\subsection{Next Category}

\begin{figure}[ht!]
    \centering
    \includegraphics[width=0.9\linewidth]{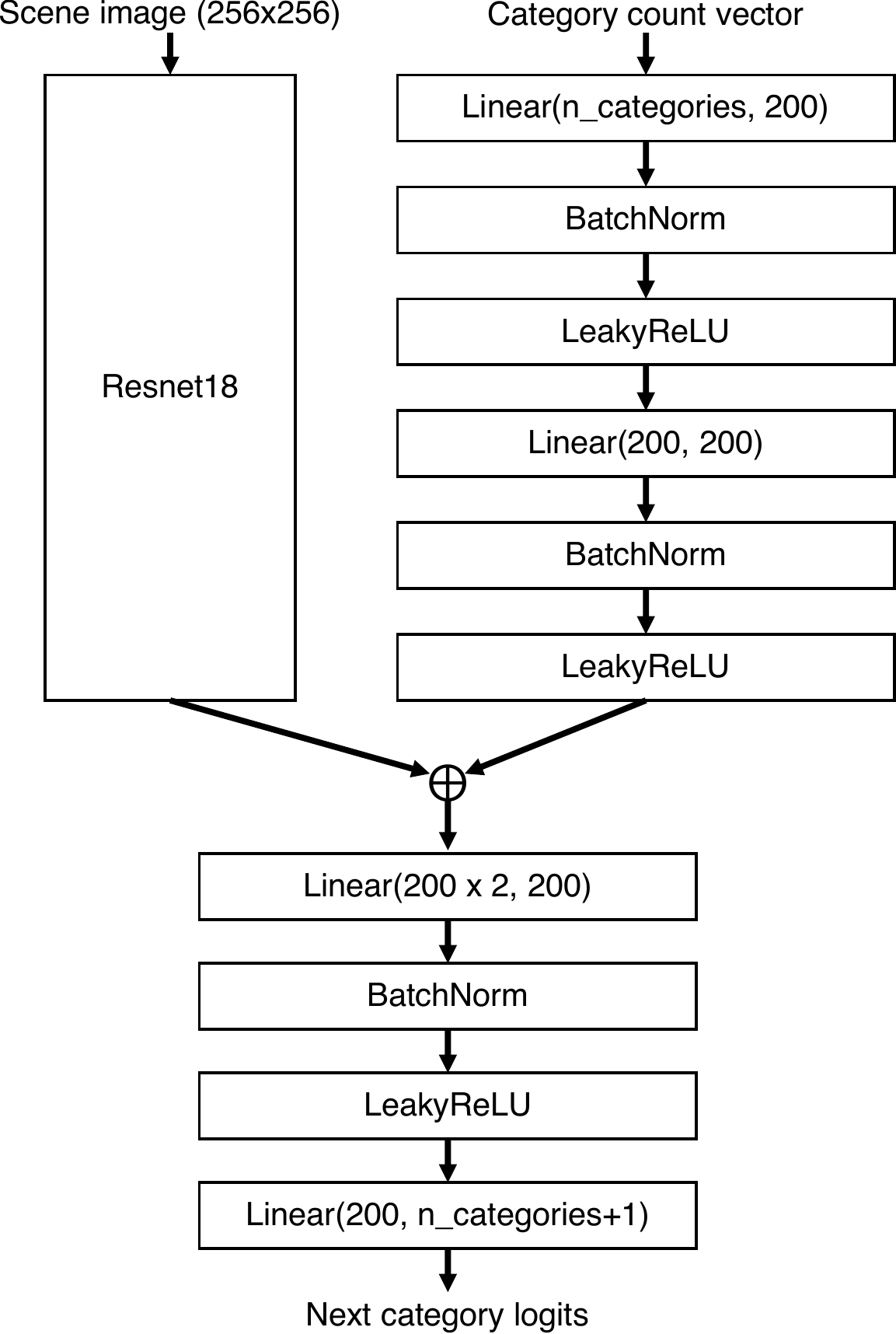}
\caption{Architecture diagram for the next category prediction module.}
\label{fig:arch_cat}
\end{figure}

The module uses a Resnet18~\cite{ResNet} to encode the scene image.
It also extract the counts of all categories of objects in the scene (i.e. a ``bag of categories'' representation), as in prior work~\cite{DeepSynthSIGGRAPH2018}, and encodes this with a fully-connected network.
Finally, the model concatenates these two encodings and feeds them through another fully-connected network to output a probability distribution over categories.
At test time, the module samples from the predicted distribution to select the next category.
Figure~\ref{fig:arch_cat} shows the architecture diagram for this network.

\subsection{Location}

\begin{figure}[ht!]
    \centering
    \includegraphics[width=0.5\linewidth]{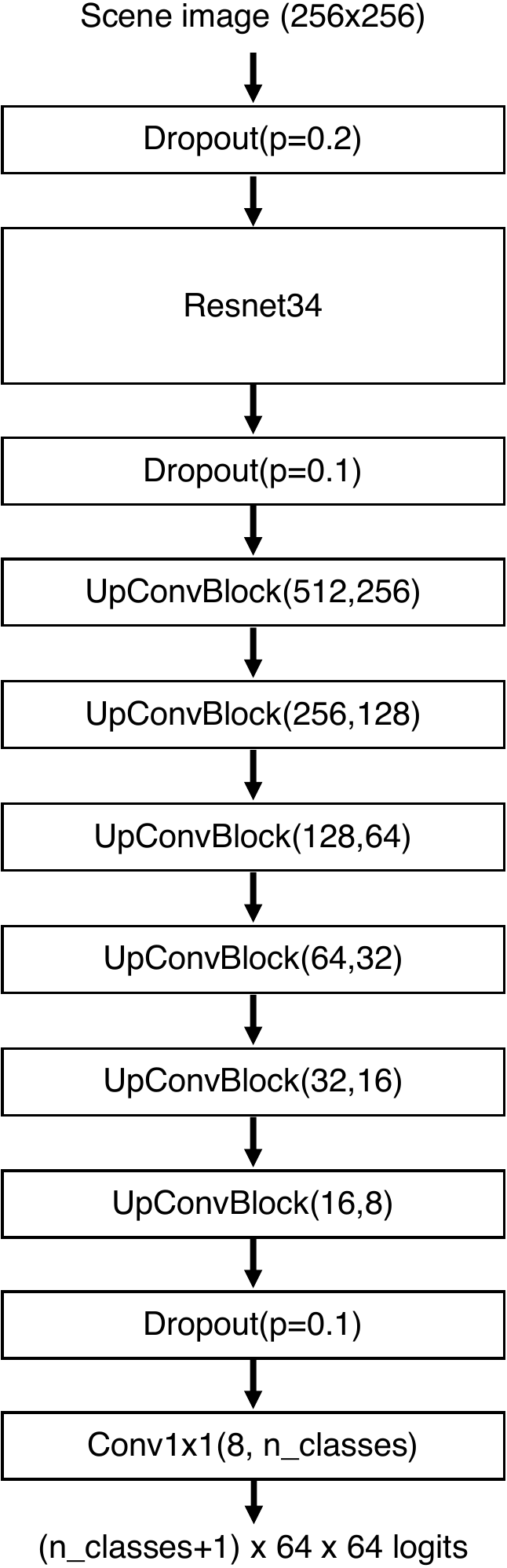}
\caption{Architecture diagram for the location prediction module. An \emph{UpConvBlock} is a 3x3 transpose convolution with stride 2 followed by a Batch Normalization layer and a ReLU layer.}
\label{fig:arch_loc}
\end{figure}

Figure~\ref{fig:arch_loc} shows the architecture diagram for this module.
It uses a Resnet34~\cite{ResNet} to encode the scene image. 
It is followed by five ``up-convolutional'' (i.e. transpose convolution) blocks (\emph{UpConvBlock}).
Up-convolution is done by first nearest-neighbor upsampling the input with scale factor of 2, and then applying a 3x3 convolution. 
Finally, we apply a 1x1 convolution to generate a $(C+1) \times 64 \times 64$ distribution over categories and location, where $C$ is the number of categories for the room type.

Since the target output during the training process (exact location of the object centroids for the room) is different from the outcome we prefer (a smooth distribution over all possible locations), the module has a high potential to overfit. 
To alleviate this, we apply dropout before and after the Resnet34 encoder, and also before the final 1x1 convolution. We also apply L2 regularization in the training process.
We found this combination of techniques effective at preventing overfitting, though we have not quantitatively evaluated the behavior of each individual component.

\subsection{Orientation}

Given a translated top-down scene image and object category, the orientation module predicts what direction an object of that category should face if placed at the center of the image.
Figure~\ref{fig:arch_orient} shows the architecture diagram for this module.
We assume each category has a canonical front-facing direction.
Rather than predict the angle of rotation $\theta$, which is circular, we instead predict the front direction vector, i.e. $[\cos \theta, \sin \theta]$.
This must be a normalized vector, i.e. the magnitude of $\sin \theta$ must be $\sqrt{1 - \cos^2 \theta}$.
Thus, our module predicts $\cos \theta$ along with a Boolean value giving the sign of $\sin \theta$ (more precisely, it predicts the probability that $\sin \theta$ is positive).
Here, we found using separate network weights per category to be most effective.

The set of possible orientations has the potential to be multimodal: for instance, a bed in the corner of a room may be backed up against either wall of the corner.
To allow our module to model this behavior, we implement it with a conditional variational autoencoder (CVAE)~\cite{CVAE}.
Specifically, we use a CNN to encode the input scene (the \emph{Conditional Prior}), which we then concatenate with a latent code $z$ sampled from a multivariate unit normal distribution, and then feed to a fully-connected \emph{Decoder} to produce $\cos \theta$ and the sign of $\sin \theta$.
At training time, we use the standard CVAE loss formulation to learn an approximate posterior distribution over latent codes).

Since interior scenes are frequently enclosed by rectilinear architecture, objects in them are often precisely aligned to cardinal directions.
A CVAE, however, being a probabilistic model, samples noisy directions.
To allow our module to produce precise alignments when appropriate, this module includes a second CNN (the \emph{Snap Predictor}) which takes the input scene and predicts whether the object to be inserted should have its predicted orientation ``snapped'' to the nearest of the four cardinal directions.

\subsection{Dimensions}

Given a scene image transformed into the local coordinate frame of a particular object category, the dimensions module predicts the spatial extent of the object.
That is, it predicts an object-space bounding box for the object to be inserted.
This is also a multimodal problem, even more so than orientation (e.g. many wardrobes of varying lengths can fit against the same wall).
Again, we use a CVAE for this: a CNN encodes the scene, concatenates it with $z$, and then uses a fully-connected decoder to produce the $[x,y]$ dimensions of the bounding box.
Figure~\ref{fig:arch_dims} shows the architecture diagram for this module.

The human eye is very sensitive to errors in size, e.g. a too-large object that penetrates the wall next to it.
To fine-tune the prediction results, we include an adversarial loss term in the CVAE training.
This loss uses a convolutional \emph{Discriminator} which takes the input scene concatenated channel-wise with the signed distance field (SDF) of the predicted bounding box.
As with orientation, this module also uses separate network weights per category.


\twocolumn[{
\renewcommand\twocolumn[1][]{#1}
\begin{center}
    \includegraphics[width=\linewidth]{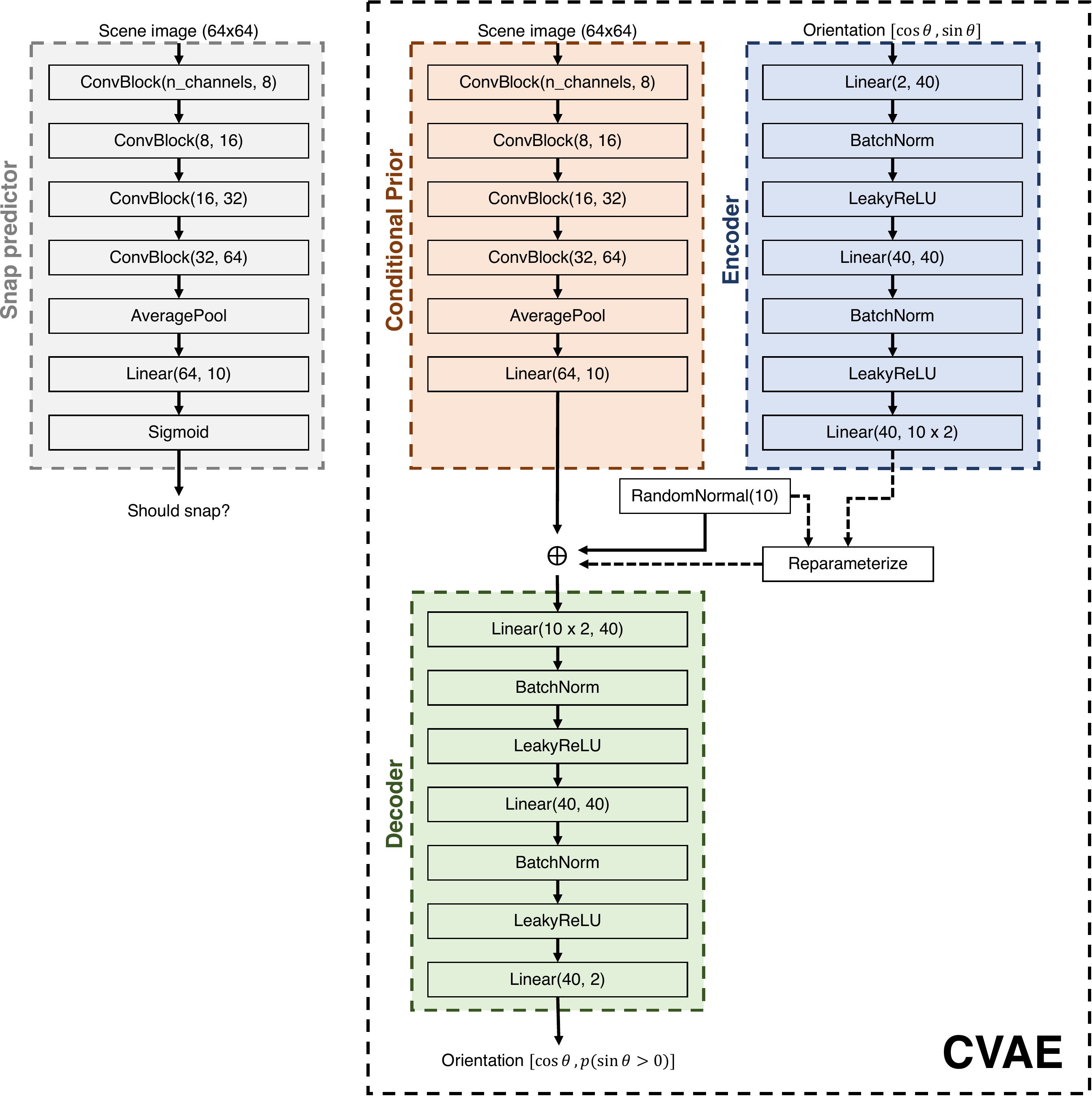}
\end{center}
\captionof{figure}{Architecture diagram for the orientation prediction module. A \emph{ConvBlock} is a 3x3 convolution with stride 2 followed by a Batch Normalization layer and a ReLU layer.}
\label{fig:arch_orient}
}]


\twocolumn[{
\renewcommand\twocolumn[1][]{#1}
\begin{center}
    \includegraphics[width=0.7\linewidth]{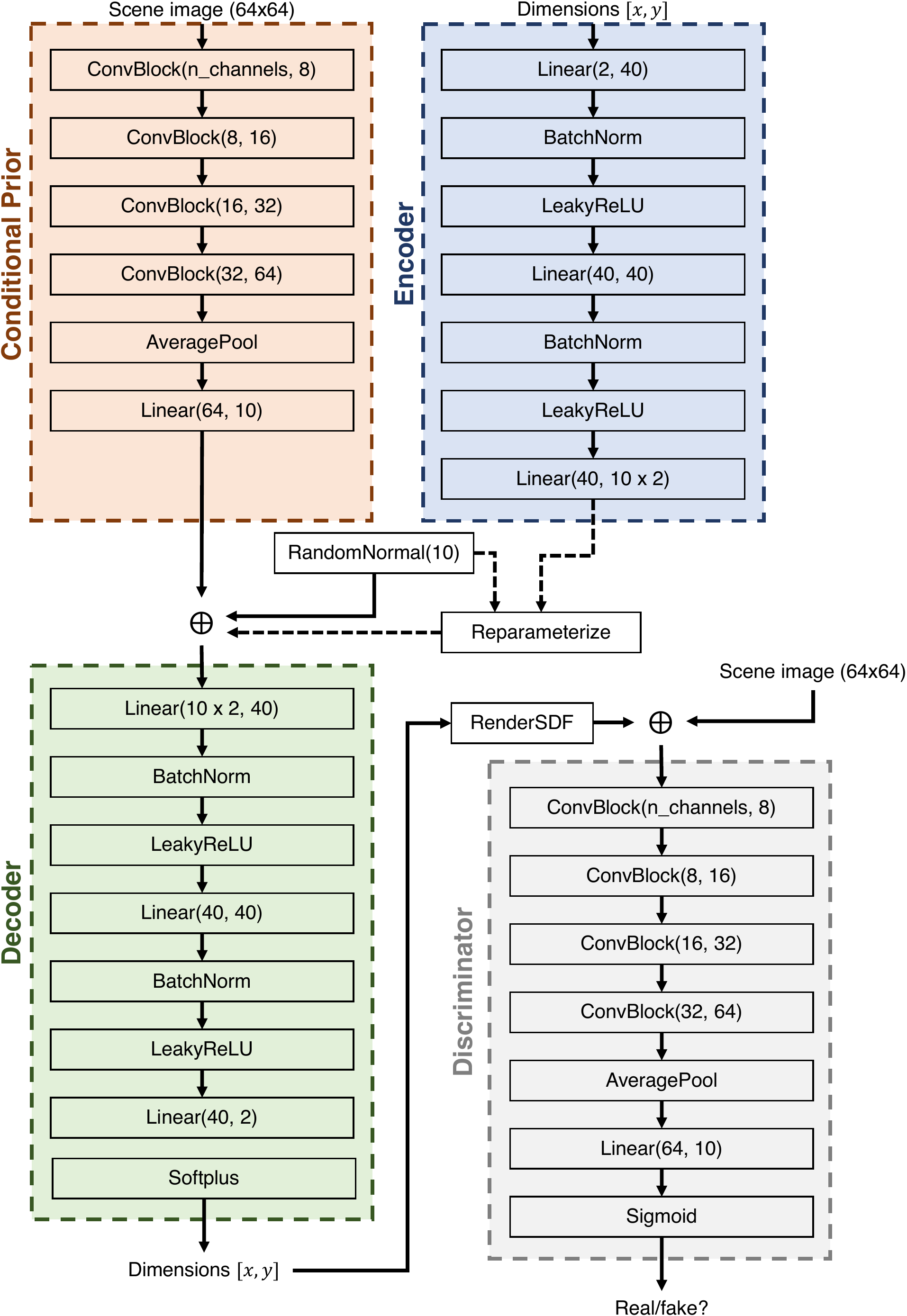}
\end{center}
\captionof{figure}{Architecture diagram for the dimensions prediction module. A \emph{ConvBlock} is a 3x3 convolution with stride 2 followed by a Batch Normalization layer and a ReLU layer.}
\label{fig:arch_dims}
}]

\cleardoublepage

\section{Dataset Details}
\label{appendix:dataset}

We adopt similar dataset filtering strategies as that of prior work~\cite{DeepSynthSIGGRAPH2018}, with a few notable differences:

\begin{enumerate}
    \item We manually selected a list of frequently-occurring objects, which we allow to appear on top of other objects (only on the visible top surface, i.e. no televisions contained in a TV stand). We remove all second tier objects whose parents were filtered out.
    \item  To facilitate matching objects by bounding box dimensions, we discard rooms containing objects which are scaled by more than $10\%$ along any dimensions. For objects scaled by less than that, we remove the scaling from their transformation matrices.
    \item We augment the living room and office dataset with 4 different rotations ($0^\circ, 90^\circ, 180^\circ, 270^\circ$) of the same room during training, to reduce overfitting, particularly for the location module. 
\end{enumerate}

Table~\ref{tab:category_counts} shows the counts of all categories appearing in the four types of rooms used in this work, where possible second tier categories are highlighted with bold.

\begin{table*}[ht!]
    \centering
    \scalebox{0.82}{
    \begin{tabular}{|c|c|c|c|c|c|c|c|c|c|}
    \hline
    &\multicolumn{9}{c|}{\textbf{Bedroom}} \\
    \hline
    \textbf{Name} & door & window & wardrobe & stand & double bed & \textbf{table lamp} & desk & dresser & office chair \\
    \textbf{Count} & 8335 & 7104 & 6690 & 6508 & 4230 & 2844 & 2586 & 2222 & 2077\\
    \hline
    \textbf{Name} & single bed & dressing table & tv stand & floor lamp & \textbf{plant} & \textbf{television} & ottoman & coffee table & \textbf{laptop}\\
    \textbf{Count} & 2032 & 1730 & 1432 & 1387 & 1341 & 1252 & 1142 & 1048 & 1010 \\
    \hline
    \textbf{Name} & shelving & \textbf{book} & sofa chair & armchair & shoes cabinet & straight chair & bunker bed & hanger & loudspeaker \\
    \textbf{Count} & 901 & 858 & 727 & 703 & 619 & 595 & 593 & 540 & 438 \\
    \hline
    \textbf{Name} & \textbf{vase} & sofa & \textbf{console} & pedestal fan & baby bed & \textbf{toy} & daybed & stool & bench chair \\
    \textbf{Count} & 395 & 393 & 368 & 320 & 266 & 231 & 181 & 178 & 134 \\
    \hline
    \textbf{Name} & whiteboard & piano & chair & \textbf{fishbowl} &&&&&\\
    \textbf{Count} & 105 & 102 & 98 & 59 &&&&&\\
    \hline
    &\multicolumn{9}{c|}{\textbf{Living Room}} \\
    \hline
    \textbf{Name} & door & window & sofa & coffee table & \textbf{plant} & sofa chair & tv stand & floor lamp & \textbf{television} \\
    \textbf{Count} & 2286 & 1789 & 1661 & 1336 & 1059 & 983 & 696 & 651 & 447 \\
    \hline
    \textbf{Name} & loudspeaker & ottoman & shelving & \textbf{vase} & fireplace & armchair & \textbf{console} & wardrobe & stand \\
    \textbf{Count} & 384 & 314 & 309 & 282 & 257 & 204 & 187 & 172 & 101 \\
    \hline
    \textbf{Name} & piano & \textbf{laptop} & \textbf{book} & \textbf{table lamp} & dresser & \textbf{cup} & \textbf{toy} & straight chair & pedestal fan \\
    \textbf{Count} & 85 & 72 & 69 & 68 & 64 & 59 & 56 & 51 & 40 \\
    \hline
    \textbf{Name} & hanger & stool & shoes cabinet & bench chair & \textbf{fishbowl} & \textbf{fruit bowl} & \textbf{glass} & \textbf{bottle} &\\
    \textbf{Count} & 34 & 33 & 31 & 26 & 21 & 17 & 16 & 15 & \\
    \hline
    &\multicolumn{9}{c|}{\textbf{Office}} \\
    \hline
    \textbf{Name} & desk & office chair & door & window & shelving & \textbf{plant} & \textbf{laptop} & \textbf{book} & wardrobe \\
    \textbf{Count} & 1616 & 1577 & 1572 & 1307 & 764 & 557 & 377 & 356 & 319 \\
    \hline
    \textbf{Name} & sofa & armchair & \textbf{table lamp} & sofa chair & straight chair & floor lamp & tv stand & coffee table & \textbf{vase} \\
    \textbf{Count} & 313 & 300 & 297 & 290 & 274 & 256 & 137 & 137 & 132 \\
    \hline
    \textbf{Name} & loudspeaker & ottoman & stand & whiteboard & dresser & \textbf{television} & piano & stool & \textbf{toy}\\
    \textbf{Count} & 113  & 112 & 102 & 100 & 84 & 84 & 75 & 56 & 52 \\
    \hline
    \textbf{Name} & hanger & pedestal fan & shoes cabinet & fireplace & bench chair & water machine & \textbf{console} & \textbf{cup} & \textbf{fishbowl} \\
    \textbf{Count} & 51 & 49 & 39 & 38 & 37 & 32 & 30 & 27 & 5 \\
    \hline
    &\multicolumn{9}{c|}{\textbf{Bathroom}} \\
    \hline
    \textbf{Name} & door & toilet & bathtub & shower & sink & window & shelving & bidet & washer \\
    \textbf{Count} & 7448 & 6437 & 5441 & 5308 & 4627 & 4299 & 3963 & 1753 & 1428 \\
    \hline
    \textbf{Name} & \textbf{plant} & wardrobe & trash can & stand & floor lamp & \textbf{toy} & hanger & ottoman & dresser \\
    \textbf{Count} & 1045 & 459 & 409 & 315 & 231 & 126 & 111 & 95 & 89 \\
    \hline
    \textbf{Name} & cabinet & \textbf{vase} & coffee table & straight chair &&&&& \\
    \textbf{Count} & 57 & 31 & 31 & 20 &&&&& \\
    \hline
    \end{tabular}
    }
    \caption{Counts for all the object categories appearing in the four types of rooms used in this paper. Bold category name indicates that this can be a second tier object}
    \label{tab:category_counts}
\end{table*}

\section{Performance of Each Model Component}

Table~\ref{tab:componentperf} shows the performance of each of our modules on a held-out test set of scene data.
Different metrics are reported for different modules, as appopriate.
We have no natural baseline to which to compare these numbers.
As an alternative, we report the improvement in performance relative to a randomly-initialized network.

\begin{table*}[ht!]
    \centering
    \begin{tabular}{|c|c|c|c|c|c|c|}
    \hline
    \textbf{Room Type} & \textbf{Cat (Top1)} & \textbf{Cat (Top5)} & \textbf{Loc (X-Ent)} & \textbf{Orient (ELBo)} & \textbf{Orient-Snap (Acc.)} & \textbf{Dims (ELBo)} \\ \hline
    \multirow{2}{*}{Bedroom}  & 0.5000 & 0.8650 & 0.0030 & 0.0899 & 0.8821 & 0.0018      \\
    & (+0.4225) & (+0.7600) & (-98.61\%) & (-54.20\%) & (+0.3821) & (-97.74\%)  \\ \hline
    \multirow{2}{*}{Living}   & 0.5375 & 0.8719 & 0.0035 & 0.1093 & 0.8902 & 0.0018      \\
    & (+0.5312) & (+0.8001) & (-98.56\%) & (-44.99\%) & (+0.3902) & (-97.79\%)  \\ \hline
    \multirow{2}{*}{Office}   & 0.5664 & 0.8948 & 0.0038 & 0.0639 & 0.9482 & 0.0015      \\
    & (+0.5525) & (+0.7629) & (-98.25\%) & (-68.03\%) & (+0.4482) & (-98.14\%)  \\ \hline
    \multirow{2}{*}{Bathroom} & 0.6180 & 0.9573 & 0.0020 & 0.0906 & 0.9419 & 0.0019      \\
    & (+0.5873) & (+0.8597) & (-85.00\%) & (-53.23\%) & (+0.4419) & (-97.64\%)  \\ \hline
    \end{tabular}
\caption{Performance of each component of our model on held-out test data. \emph{Acc.} is binary classification accuracy; \emph{Top N} is top-n multiclass classification accuracy; \emph{X-Ent} is cross-entropy; \emph{ELBo} is the standard Evidence Lower Bound objective for variational autoencoders. The numbers in parentheses show the improvement relative to a randomly-initialized network.}
    \label{tab:componentperf}
\end{table*}


\section{Generalization}
\begin{figure}[ht!]
    \centering
    \includegraphics[width=0.99\linewidth]{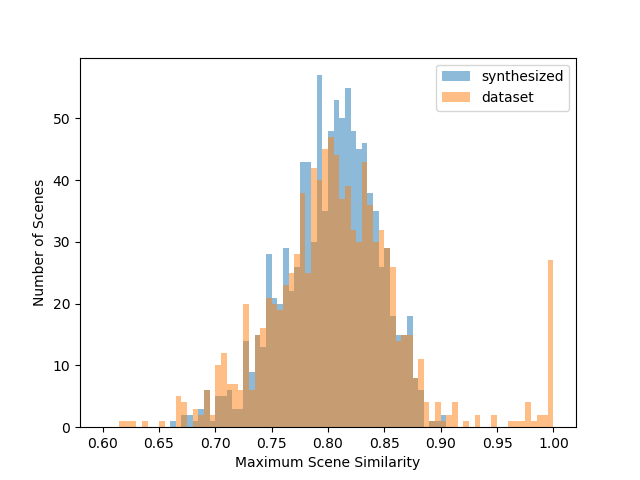}
\caption{Plotting the maximal similarity score of a bedroom against $5,000$ rooms from the training set. We plot the distribution of results for $1,000$ synthesized rooms and $1,000$ held out rooms in the training set (disjoint from the $5,000$)}.
\label{fig:generalization}
\end{figure}

To evaluate if our models are merely ``memorizing" the training scenes, we measure similar a generated room can be to a room in the training set. 
To do so, we use the same scene-to-scene similarity function as in prior work~\cite{DeepSynthSIGGRAPH2018} and compute the maximal similarity score of a generated room against $5,000$ rooms in the training set, We plot the score distribution for $1,000$ generated rooms in Figure~\ref{fig:generalization}.
For comparison, We also compute the same score for $1,000$ rooms from the training set (which are disjoint from the $5,000$ aforementioned rooms). 
In general, the behavior for the synthesized rooms is similar to that of scenes from the dataset.
Our model definitively does not just memorize the training data, as it is actually less likely for our model to synthesize a room that is very similar to one from the training set. 
It is also less likely for our model to synthesize something that is very different from all other rooms in the training set.
This is coherent with our impression: that our model suffers from minor mode collapses, and does not capture all possible unique room layouts.
Finally, the large spike in extremely-similar rooms for the dataset-to-dataset comparison (the tall orange bar on the far right of the plot) is due to exact duplicate scenes with exist in the training set.

\end{document}